\documentclass[1p, review]{elsarticle}

\usepackage{amsmath}
\usepackage{amssymb}

\usepackage{lineno}
\usepackage{hyperref}
\usepackage{algorithm}
\usepackage{algorithmic}
\usepackage[figuresright]{rotating}
\usepackage{subfigure}
\usepackage{multirow}
\usepackage[capitalise, noabbrev]{cleveref}

\usepackage{xcolor}

\usepackage{caption}
\usepackage{flushend}

\usepackage{booktabs}
\usepackage{array}
\usepackage{makecell}
\usepackage{graphbox}

\usepackage{ulem}
\usepackage{soul}

\newcolumntype{P}[1]{>{\centering\arraybackslash}p{#1}}
\newcolumntype{M}[1]{>{\centering\arraybackslash}m{#1}}

\journal{Engineering Applications of Artificial Intelligence}

\begin{document}
\begin{frontmatter}






\title{ContextMix: A context-aware data augmentation method for industrial visual inspection systems}


\author[KAIST,SEMCO]{Hyungmin Kim}
\author[KAIST]{Donghun Kim}
\author[LGAI]{Pyunghwan Ahn}
\author[DFKI,RPTU]{Sungho Suh}
\author[SEMCO]{Hansang Cho}
\author[KAIST]{Junmo Kim}

\address[KAIST]{School of Electrical Engineering, Korea Advanced Institute of Science and Technology (KAIST), Daejeon 34141, South Korea}
\address[SEMCO]{Samsung Electro-Mechanics, Suwon 16674, South Korea}
\address[LGAI]{LG AI Research, 30, Magokjungang 10-ro, Gangseo-gu, Seoul, South Korea}
\address[DFKI]{German Research Center for Artificial Intelligence (DFKI), 67663 Kaiserslautern, Germany}
\address[RPTU]{Department of Computer Science, RPTU Kaiserslautern-Landau, Kaiserslautern, Germany}

\begin{abstract}
While deep neural networks have achieved remarkable performance, data augmentation has emerged as a crucial strategy to mitigate overfitting and enhance network performance. These techniques hold particular significance in industrial manufacturing contexts. Recently, image mixing-based methods have been introduced, exhibiting improved performance on public benchmark datasets. However, their application to industrial tasks remains challenging.
The manufacturing environment generates massive amounts of unlabeled data on a daily basis, with only a few instances of abnormal data occurrences. This leads to severe data imbalance. Thus, creating well-balanced datasets is not straightforward due to the high costs associated with labeling. Nonetheless, this is a crucial step for enhancing productivity.
For this reason, we introduce ContextMix, a method tailored for industrial applications and benchmark datasets. ContextMix generates novel data by resizing entire images and integrating them into other images within the batch. This approach enables our method to learn discriminative features based on varying sizes from resized images and train informative secondary features for object recognition using occluded images.
With the minimal additional computation cost of image resizing, ContextMix enhances performance compared to existing augmentation techniques. We evaluate its effectiveness across classification, detection, and segmentation tasks using various network architectures on public benchmark datasets. Our proposed method demonstrates improved results across a range of robustness tasks. Its efficacy in real industrial environments is particularly noteworthy, as demonstrated using the passive component dataset. Our code is made available at \url{https://github.com/Hy2MK/ContextMix}.
\end{abstract}


\end{frontmatter}


\section{Introduction} \label{sec:introduction}
Modern industrial manufacturing environments have been utilizing various machine-learning techniques to increase the efficiency of production procedures. In particular, the techniques using image processing, such as those proposed by \cite{GUERRA1997365}, \cite{LYU20095113}, and \cite{REJC201110665} apply to detect defects of goods or parts, reduce human error, and improve the quality of products.
Deep neural networks (DNNs) have shown incomparable performances on various computer vision tasks, which are image classification~\cite{he2016deep, han2017deep, xie2017aggregated}, object detection~\cite{ren2015faster, liu2016ssd}, and semantic segmentation~\cite{long2015fully, Chen_2018_ECCV, zhang2020causal}.
Then, various approaches have been proposed to improve the model performance by designing more complex but efficient network architectures or gathering more large-scale datasets such as ImageNet~\cite{krizhevsky2012imagenet}, YFCC-100M~\cite{2812802}, and IG-1B-Targeted~\cite{yalniz2019billion}. In this regard, various techniques\cite{8443132, 9164986} have adopted the DNNs to visual inspection systems in the manufacturing process to improve productivity.
\begin{figure}[t]
    \centering
    \resizebox{1.\linewidth}{!}{
    \setlength{\tabcolsep}{5pt}
    \begin{tabular}{ccc}
        \includegraphics[width=0.85\linewidth]{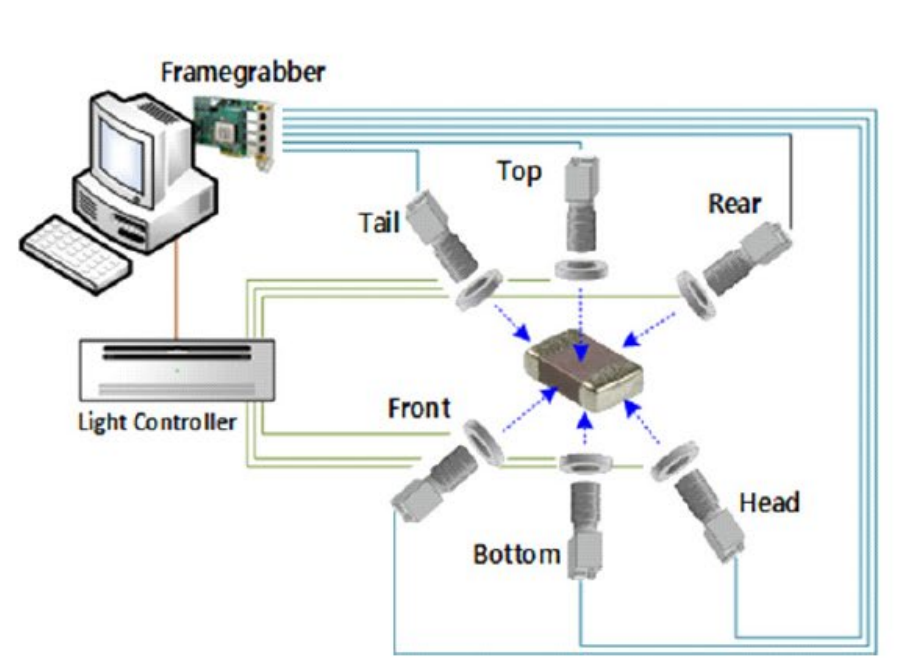}
        &&\includegraphics[width=0.85\linewidth]{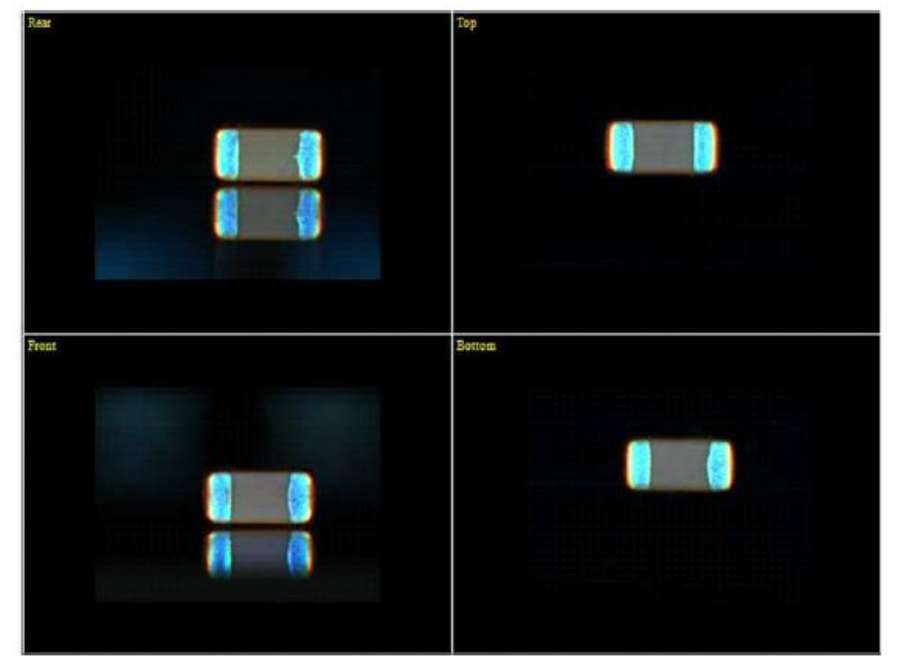}\\
        \fontsize{0.5cm}{0.5cm}\selectfont{(a) Overview of visiual inspection machine}
        &&\fontsize{0.5cm}{0.5cm}\selectfont{(b) Grabbed images}\\
    \end{tabular}}
    \caption{The machines and images are utilized by Samsung Electro-mechanics for vision inspections, and reported in~\cite{suh2017automatic}}
    \label{fig:semmachines}
\end{figure}

The usage of deeper networks with heavy parameters and increasing training dataset size may intrinsically incur the overfitting problem. In particular, the problem is more vulnerable to the industry due to the characteristics of the environment, which is available to generate massive data daily from manufacturing machines. Regularization techniques employ effective training strategies to enhance the performance of various tasks by alleviating overfitting issues, and data augmentation methods are the proper techniques for industrial applications. 
Since most of the industrial data, such as the images collected by high-speed vision inspection machines, as shown in \cref{fig:semmachines}, the images are generally unlabeled, and then data labeling and verification of the labels are essential works but require expensive costs.
In contrast, data augmentations are cost-effective since they utilize the existing dataset as possible to create unseen data.
In addition, it is rare to collect abnormal data in the manufacturing process. Most of the collected data are severely imbalanced, biased toward the normal data, and have long-tailed distribution. Rather, the severely imbalanced data hinder to train models and lead to decreasing productivity. Thus, data collection requires sufficient time and costs for well-balanced datasets. For these reasons, data augmentation methods in industrial applications have been proposed by the generative model~\cite{YUN2020317, Sebastian2021Synthetic} and by simple image processing techniques, such as image transformation and distortion~\cite{Ferguson2018Detection, LOPEZDELAROSA2022117731}. These approaches are considered suitable for their specific tasks, but could not employ in various tasks and benchmark datasets.
Additionally, there are addresses to exploit through generative-based approaches~\cite{mirza2014conditional, ZHENG20201009}, but they are unsuitable for the industrial environment. In general, the industrial environment suffers from inherent data imbalance, and there is often an insufficient quantity of data for several classes within the dataset. In such situations, generative adversarial networks(GANs)-based approaches require sufficient training data so that in industrial scenarios, they are easy to reach unstable training and model collapse. Furthermore, even after successfully training GAN model, the generated images would need to be reviewed by managers or experts to select appropriate ones. Then it means that required additional tasks and costs for them.

Recently, regional dropout and mixed-image-generating techniques have been proposed on public benchmark datasets and presented remarkably improved performances. Cutout~\cite{devries2017improved} is a symbolic method of regional dropout that erases a random region in images and then inputs the image for training, and Mixup~\cite{zhang2017mixup} generates new data to combine the pixel intensity of the two samples using arbitrary rates.
Combining Mixup and Cutout, CutMix~\cite{yun2019cutmix} has recorded the best performances on various tasks among regional dropout methods. In this regard, several industrial applications~\cite{SHIN2022107996, QI2022116473} with the above methods have been proposed and shown improvement in performance. However, the above methods have intrinsic drawbacks. In particular, CutMix utilizes an uninformative region by cropping the background, and it becomes even worse when they are pasted over informative regions of the other image. CutMix variants have been proposed to address this problem by avoiding cutting out the important regions from images. PuzzleMix~\cite{kim2020puzzle} identifies object regions using saliency information~\cite{simonyan2013deep, wang2015deep} and generates new mixed images by preserving the discriminative parts. These saliency-based methods are helpful tools to understand the essential region of images for object recognition but have a few shortcomings. Firstly, the methods required incurring a non-negligible increase in computational costs for inferring saliency information. Second, we contemplate the integrity of salient information. In general, the salient regions are object regions. The problem is that the tools do not guarantee that the salient regions always indicate object regions precisely. The highlighted regions by saliency information are the results of networks trained, and networks tend to concentrate on learning uncomplicated but discriminative features. For this reason, the networks are possible to treat backgrounds are more critical information than the actual objects. The worst case is mixing only background images that are highlighted on saliency maps.
Then, there is no object information, and the network should train using only background images and classify what class is. In this regard, the images do not match the saliency-based methods scheme, rather are increasing ambiguity for network training. Moreover, well-extracted object regions by saliency maps are insufficient for object recognition tasks. From the human perspective, when we recognize objects, we simultaneously consider what they are and where they are.
\begin{figure}[t]
    \centering
    \resizebox{0.9\linewidth}{!}{
    \rotatebox{90}{
    \begin{tabular}{ccc}
        \fontsize{15.cm}{15.cm}\selectfont\textbf{ContextMix}
        &\fontsize{15.cm}{15.cm}\selectfont\textbf{PuzzleMix}
        &\fontsize{15.cm}{15.cm}\selectfont\textbf{CutMix}\\
        &&\\&&\\&&\\
        \rotatebox{270}{\includegraphics[width=0.6\linewidth]{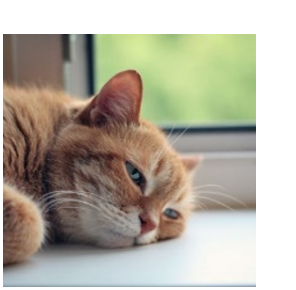}}
        &\rotatebox{270}{\includegraphics[width=0.6\linewidth]{figure/img27_1_ours.png}}
        &\rotatebox{270}{\includegraphics[width=0.6\linewidth]{figure/img27_1_ours.png}}\\
        &&\\&&\\&&\\
        \rotatebox{270}{\includegraphics[width=0.6\linewidth]{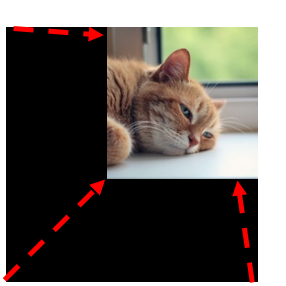}}
        &\rotatebox{270}{\includegraphics[width=0.66\linewidth]{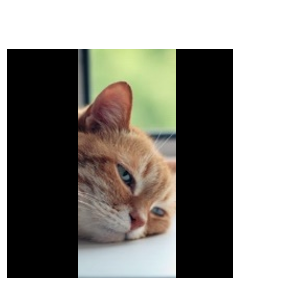}}
        &\rotatebox{270}{\includegraphics[width=0.6\linewidth]{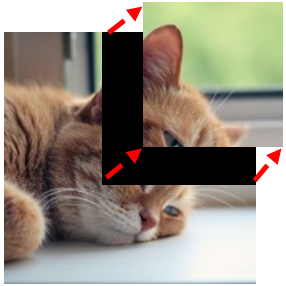}}\\
        &&\\&&\\&&\\
        \rotatebox{270}{\includegraphics[width=0.6\linewidth]{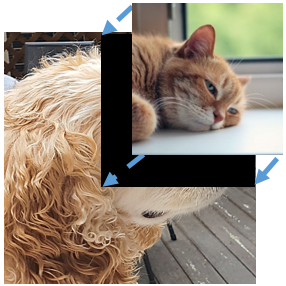}}
        &\rotatebox{270}{\includegraphics[width=0.66\linewidth]{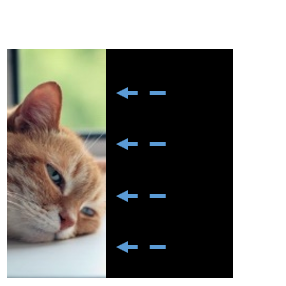}}
        &\rotatebox{270}{\includegraphics[width=0.6\linewidth]{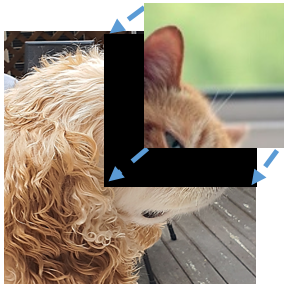}}\\
        &&\\&&\\&&\\
        \rotatebox{270}{\includegraphics[width=0.6\linewidth]{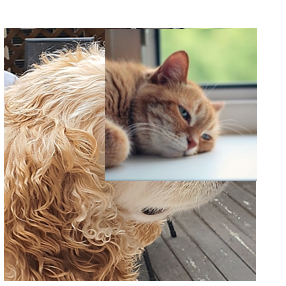}}
        &\rotatebox{270}{\includegraphics[width=0.66\linewidth]{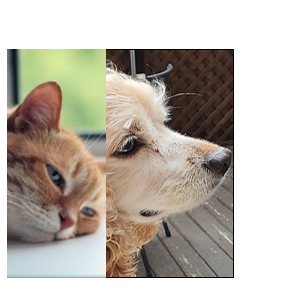}}
        &\rotatebox{270}{\includegraphics[width=0.6\linewidth]{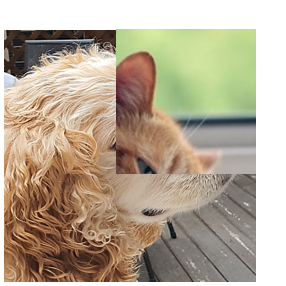}}\\
    \end{tabular}}}
    \caption{Overview of CutMix \cite{yun2019cutmix}, PuzzleMix \cite{kim2020puzzle}, and ContextMix.}    
    \label{fig:overview}
\end{figure}
For example, if someone sees a polar bear sitting on a couch in the house, the observer would probably have a lot of questions about what it is or why the bear is sitting there. It means that an object should be in the right place in a common sense so that people can recognize it easily without a doubt. The context-aware network~\cite{Dvornik2018ECCV} and the context generative model~\cite{Tripathi_2019_CVPR} verify that randomly pasting objects on images leads to degraded object recognition performance unless the object is located in the proper context. These methods show the importance of scene understanding for recognizing objects. Then they claim that context is also important information as object information, and the model should learn both the object and the contextual information.

To address and overcome these problems, we propose a simple and intuitive method called ContextMix.
As shown in \cref{fig:overview}, we paste the entire image to the occluded region by resizing.
The labels are combined proportionally to the area of the occluded region and the remaining region. A sample generated by ContextMix contains two important features: information from the resized whole image and information from the occluded image of the original size. Resized image preserves the whole image structure which has the object and the contextual information. The occluded image of the original size has partial information about the object. 
ContextMix provides more opportunities to learn discriminative object and context information, including background, at the negligible additional computation cost of image resizing.
To show the effectiveness of the proposed method, we evaluate the proposed method with different network architectures on various tasks and datasets, including the industrial dataset.

The main contributions of the proposed method can be summarized as follows.
\begin{itemize}
    \item We propose a novel and effective image data augmentation method, named ContextMix. The proposed method does not require the saliency regions, so negligible computation costs are needed to compare other methods.
    \item To alleviate the overfitting issue, we experimentally analyze through diverse ablations to confirm the effectiveness of methods that require not only focusing on the saliency regions but also considering contextual information.
    \item We demonstrate that ContextMix improves performance through various benchmark datasets, network architectures, and tasks. In particular, we confirm the effectiveness of the proposed method on the real-world industrial dataset.
\end{itemize}

The rest of the paper is organized as follows. \cref{sec:related_work} introduces related work. \cref{sec:method} provides the details of the proposed method. \cref{sec:exps} presents qualitative and quantitative experimental results with a variety of network architectures on multiple datasets. In \cref{sec:discussion}, we analyze the crucial factors of our proposed method using various studies and confirm the impacts of random resized images. At last, \cref{sec:conclusion} concludes the paper.

\section{Related work} \label{sec:related_work}
\subsection{Regional Dropout}
The regional dropout has exhibited improved generalization by training models using the most discriminative parts of objects. Cutout~\cite{devries2017improved} is a representative regional dropout method that removes a random square region in images sampled using a random function. Hide and Seek (HaS)~\cite{singh2017hide} is a method motivated by Cutout that removes multiple random regions.
Mixup~\cite{zhang2017mixup} generates additional data from existing training data by combining pixel colors of the samples with an arbitrary mixing rate. While the sample generated by Mixup has a blurry and imprecise image structure, features of each sample in a mixed form ratio can still be retained. CutMix~\cite{yun2019cutmix} proposed a method combining Mixup and Cutout. In CutMix, new samples are generated by replacing patches of random regions by crops from other images.
However, this simple and effective method has several intrinsic shortcomings, which are cropping and pasting background images, not objects.

To overcome the drawbacks of CutMix, the method variants have been proposed diversely. One of the attempts is to modify the shapes of the masks. CowMask~\cite{french2020milking} generates an atypical mask from Gaussian noise to combine two images, and FMix~\cite{Harris2020fmix} obtains arbitrary masks from Fourier-transformed images. However, it still suffers from the possibility of cropping uninformative regions of the image. To avoid this, class activation map (CAM)~\cite{zhou2016learning} or saliency information~\cite{simonyan2013deep, wang2015deep} are used to perceive the informative object regions before cropping and pasting. Attentive CutMix~\cite{walawalkar2020attentive} and SaliencyMix~\cite{uddin2021saliencymix} methods focused on preserving discriminative features for object recognition on the cropped region using CAM and saliency maps. While PuzzleMix~\cite{kim2020puzzle} and Co-Mixup~\cite{kim2020co} adopted an optimal transportation concept and saliency information to preserve the important regions for both images that are mixed. These methods can avoid the case where the overridden regions contain valuable information about images by relocating the regions in an optimal manner. However, the additional computational cost for the method is required to obtain saliency information from an extra forward pass through the network. Furthermore, the salient regions are not always the object. The highlighted regions are possible in the background. In this case, the saliency-based methods lead to generating images that have background only without object information. For network training, such images are harmful. More recently, StyleMix-based methods \cite{Hong2021StyleMixSC}, \cite{zhou2021mixstyle} proposed to adopt the style transfer scheme to generate new images by mixing the style of images through the sub-network presented in adaptive instance normalization (AdaIN) \cite{huang2017arbitrary}. However, the method requires pre-trained style transfer networks, and the generated images' quality depends on the pre-trained networks.

On the other hand, the cropped regions do not perpetually harm performance but rather give more chances to learn diverse features of objects to recognize. Adversarial Erasing (AE) improved localization \cite{zhang2018adversarial, Kim_2017_ICCV} and segmentation \cite{wei2017object} based on regional dropout. The methods erase salient regions on input data that are small but discriminative, then erased data is re-input for finding the following salient regions. In this fashion, the methods expanded object regions progressively. It means occluded regions do not affect losing information, crucially; instead, it helps to learn features overall, not only a few discriminative focused.

Also, the regional dropout method is a representative regularization method on the input data level. For more efficient training networks, the method can be trained efficiently without modification of the network architectures and should have compatibility with other regularization methods, such as Dropout \cite{srivastava2014dropout}, Batch normalization \cite{ioffe2015batch}, and ShakeDrop \cite{yamada2019shakedrop}. ContextMix operates on the input layer following the regional dropout scheme, and we demonstrate the compatibility of our proposed method with other regularization methods.

\subsection{Visual Context}
The importance of visual contextual information for recognizing objects is shown by context-aware networks \cite{Dvornik2018ECCV, 8941244} and generative model \cite{Tripathi_2019_CVPR}. The methods argued that mixing objects on images without considering contextual information degrade performance. By understanding scenes, the methods locate the suitable place to paste objects on images for generating natural-looking images, and it led to improving performances on detection and segmentation tasks. Recently proposed mixing data augmentation methods focused on discriminating object regions from the background images using additional inferences such as saliency information, not considering contextual information. From the visual context methods, classification tasks also should consider adjusting contextual information to improve performance.

\subsection{Industrial applications using data augmentation}
Various data augmentation techniques have been utilized in industrial applications.
Geometric image transformations and image distortions or combinations of the methods are the most widespread since these are easy to apply their applications and preserve the image characteristics of their products. \cite{LOPEZDELAROSA2022117731} utilized only geometric image transformations to classify defects in the sensor images. \cite{Ferguson2018Detection} proposed to detect defective casting or welding in the X-ray images, using image flips, Gaussian blur, and cropping. \cite{8354135} and \cite{BERGS2020947} addressed the utilization of image rotation, flipping, and distortion for wafer defects and cutting tool wear detection. Another approach is the utilizing generative models in the manufacturing industry to overcome insufficient training data that abnormal or defective data occurred rarely. \cite{YUN2020317} proposed to improve defect classification performance using generated images by the conditional variational autoencoder.
\cite{Sebastian2021Synthetic, 10058512, 9471877, CHEN2022126580, met12020311, 10.1007/s00521-021-05982-z} proposed to generate images using the to acquire enough quantity of the training dataset. Furthermore, \cite{9329076} addressed adopting the auxiliary guide loss for the generator model and improved the quality of defect data and diagnostic accuracy of their production process.
Nonetheless, these generated data should undergo a verification process by managers and experts familiar with the manufacturing environment and possessing domain knowledge to ensure data clarity.
Several studies proposed~\cite{SHIN2022107996, 9101484, 9250552, 9631048} using Mixup method for their targets, which are detecting defective objects on wafers, sensing fault diagnosis of chemical product processes, and capturing signals for motion recognition. 
However, \citep{ZHOU2022117351, QI2022116473} addressed that the image mixing-based approaches are available to harm their image or signal characteristics, so utilizing the methods for each product should be considered.

\section{Methodology} \label{sec:method}
\subsection{Principal approach}
Our proposed method is a regional dropout-based data augmentation approach to address how to generate new images and new labels of the images using two input images and each label.
Thus we define input images and labels first.
Let the clean input image be as $\mathit{x}$,
and the label of $\mathit{x}$ be as $\mathit{y}$. The newly generated image and its label are denoted as $\mathit{(\tilde{x}, \tilde{y})}$. The image $\mathit{\tilde{x}}$ is created by mixing two training images and their labels, denoted by $(x_A, y_A)$ and $(x_B, y_B)$, respectively.
The combining operation is defined from inspired by CutMix \cite{yun2019cutmix} as:
\begin{linenomath}
\begin{equation}
\begin{aligned}
    &\tilde{x} = M\odot{x_A} + (1-M)\odot{x_B}\\
    &\tilde{y} = \lambda\cdot y_A + (1-\lambda)\cdot y_B
\end{aligned}
\label{eq:1}
\end{equation}
\end{linenomath}
\noindent where $\mathit{M}$ denotes binary masks for cutting out region from image $\mathit{x_A}$ and $\odot{}$ denotes element-wise multiplication.
The bounding box coordinates $\mathit{B(r_{xs},r_{ys},r_{xe},r_{ye})}$, where $\mathit{(r_{xs},r_{ys})}$ denotes the top left corner and $\mathit{(r_{xe},r_{ye})}$ denotes the bottom right corner, are obtained by sampling $\mathit{(r_{xs},r_{ys})}$ from the uniform distribution and adjusting the area of the box according to the ratio of areas, $\mathit{\lambda}$.
\begin{equation}
    \lambda = 1 - (w \times h)/(W \times H)
\label{eq:2}
\end{equation}
\noindent where $\mathit{w = r_{xe}-r_{xs}}$ is the width of the cropped area and $\mathit{h = r_{ye}-r_{ys}}$ is the height, and W, H are width and height of the original image. Thus, the cropped size can be represented as $\mathit{w\times h}$. The new label $\mathit{\tilde{y}}$ is calculated by combining $\mathit{y_A}$ and $\mathit{y_B}$ proportionally to the areas of the regions occupied by each image.

We propose to paste the resized whole image of $\mathit{x_B}$ to $\mathit{x_A}$ so that the generated new image does not have an obscured partial image of $\mathit{x_B}$. Rather, it preserves the entire image structure, including object and context, through resizing $\mathit{x_B}$.
The new label $\mathit{\tilde{y}}$ is determined by the resize ratio properties, which are defined as follows:
\begin{equation}
\begin{aligned}
    &\tilde{y} = \lambda_{A}\cdot\epsilon_{A}\cdot y_A + \lambda_{B}\cdot\epsilon_{B}\cdot y_B\\
    &\lambda_{A} = 1 - (w \times h)/(W \times H)\\
    &\epsilon_{A} = (W \times H)/(W \times H) = 1\\
    &\lambda_{B} = (w \times h)/(W^{\prime} \times H^{\prime})\\
    &\epsilon_{B} = (W^{\prime} \times H^{\prime})/(W \times H)
\end{aligned}
\label{eq:3}
\end{equation}
\begin{figure}[t]
    \centering
    \resizebox{0.9\linewidth}{!}{
    \setlength{\tabcolsep}{5pt}
    \begin{tabular}{cccccc}
        \includegraphics[width=0.7\linewidth]{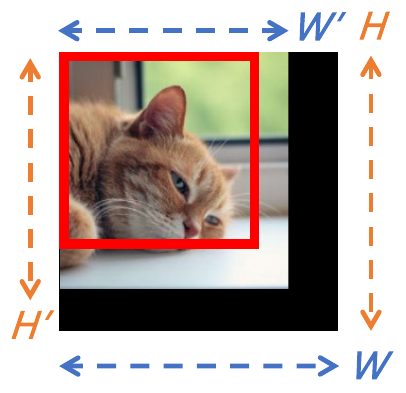} &\quad\quad\quad\quad\quad\quad\quad\quad &\quad\quad\quad\quad\quad\quad\quad\quad &\quad\quad\quad\quad\quad\quad\quad\quad &\quad\quad\quad\quad\quad\quad\quad\quad &\includegraphics[width=0.7\linewidth]{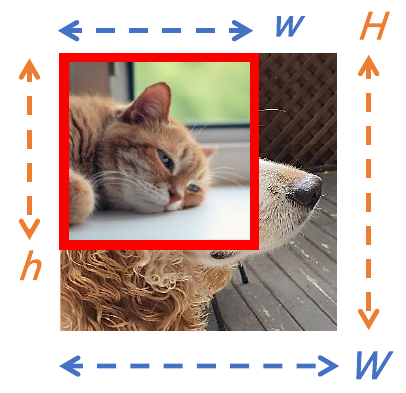}\\
        \fontsize{0.7cm}{0.7cm}\selectfont{(a) Resized Image~$x^{\prime}$} &\quad\quad\quad\quad\quad\quad\quad\quad &\quad\quad\quad\quad\quad\quad\quad\quad &\quad\quad\quad\quad\quad\quad\quad\quad &\quad\quad\quad\quad\quad\quad\quad\quad &\fontsize{0.7cm}{0.7cm}\selectfont{(b) Generated New Image~$\tilde{x}$}\\
    \end{tabular}}
    \caption{Resize ratio: $W$ and $H$ are the original image size, $w$ and $h$ are the cropped size. The image $x^{\prime}$ is resized as $\mathit{W^{\prime}}$ and $\mathit{H^{\prime}}$. Resize rate, $\mathit{\epsilon}$ is defined as $(W^{\prime}\times H^{\prime})/(W\times H)$. For ContextMix, $\mathit{W^{\prime}}$ is set to $\mathit{w}$ and $\mathit{H^{\prime}}$ is set to $\mathit{h}$.}
    \label{fig:resize_ratio}
\end{figure}
\noindent where $\mathit{W\times H}$ is the original image size, and $\mathit{W^{\prime} \times H^{\prime}}$ is the resized image size as shown in \cref{fig:resize_ratio}.
We then define $\mathit{\epsilon}$ as the resize ratio. $\mathit{x_A}$ maintains the original ratio, so $\mathit{\epsilon_{A}}$ is 1. $\mathit{x_B}$ is resized to $\mathit{x_B^{\prime}}$, and $\mathit{\epsilon_{B}}$ is $(W^{\prime} \times H^{\prime})/(W \times H)$. We also define $\mathit{\lambda}$ as the ratio of the area occupied by each image. $\mathit{\lambda_{A}}$ is equivalent to $\mathit{\lambda}$ in CutMix, while $\mathit{\lambda_{B}}$ is the proportion of area occupied by the resized image $\mathit{x_B^{\prime}}$. For ContextMix, the resized image fits cropped region size. Therefore, $\mathit{W^{\prime}\times H^{\prime}}$ should be equal to $\mathit{w\times h}$. In other words, we set $\mathit{W^{\prime}}$ to $\mathit{w}$ and $\mathit{H^{\prime}}$ to $\mathit{h}$. 
To sum up, in a mini-batch, ContextMix-ed new image data $\mathit{\tilde{x}}$ and labels $\mathit{\tilde{y}}$ are generated according to the following equation:
\begin{equation}
\begin{aligned}
    &\tilde{x} = M\odot{x_A} + (1-M)\odot{x_B^{\prime}}\\
    &\tilde{y} = \lambda_{A}\cdot 1\cdot y_{A} + \lambda_{B}\cdot \epsilon_{B}\cdot y_{B}
\end{aligned}
\label{eq:4}
\end{equation}

\cref{algo:algo1} presents the code-level description of ContextMix algorithm that explains how to get the mixed images and labels in an epoch.
\begin{algorithm}[t]
    \centering
	\caption{Procedure of ContextMix}
	\label{algo:algo1}
	\begin{algorithmic}[1]
		\FOR {$iteration=1,2,\ldots$}
		    \STATE inputs, targets = get$\_$minibatch(loader)
		    \STATE
		    \STATE $\mathit{W}$, $\mathit{H}$ = calculate$\_$image$\_$size(inputs)
		    \STATE $\mathit{y_A}, \mathit{y_B}$ = targets, shuffle$\_$index(targets)
		    \STATE $\mathit{x_{center}, y_{center} = }$~calculate\_crop\_center\_position$\mathit(W, H)$
		    \STATE $\mathit{r_{xs}, r_{ys}, r_{xe}, r_{ye} = }$~calculate$\_$crop$\_$region$\mathit(x_{center}, y_{center}, W, H)$
		    \STATE $\mathit{w, h =}$~calculate$\_$adjusted$\_$crop$\_$image$\_$size$\mathit{(r_{xs}, r_{ys}, r_{xe}, r_{ye})}$
		    \STATE \textbf{inputs$^{\prime}$ = resize$\_$images(inputs, $\mathit{w, h}$)}
		    \STATE \textbf{inputs[~:~,~:~,~$r_{xs}$:$r_{xe}$~,~$r_{ys}$:$r_{ye}$~] = inputs$^{\prime}$}
            \STATE
			\STATE $\lambda_A = 1-(w\times h)/(W\times H)$
			\STATE $\lambda_B = (w\times h)/(w\times h) = 1$
			\STATE $\epsilon_A = (W\times H)/(W\times H) = 1$
			\STATE $\epsilon_B = (w\times h)/(W\times H)$
			\STATE \textbf{targets = $\lambda_A\cdot\epsilon_A\cdot y_{A} + \lambda_B\cdot\epsilon_B\cdot y_{B}$}
            \STATE
            \STATE outputs = model(inputs)
            \STATE loss = calculate$\_$loss(outputs, targets)
            \STATE update$\_$model(loss)
		\ENDFOR
	\end{algorithmic}
\end{algorithm}
\begin{figure}[t]
    \centering
    \resizebox{1.\linewidth}{!}{
    \setlength{\tabcolsep}{5pt}
    \begin{tabular}{c}
        \includegraphics[width=0.9\linewidth]{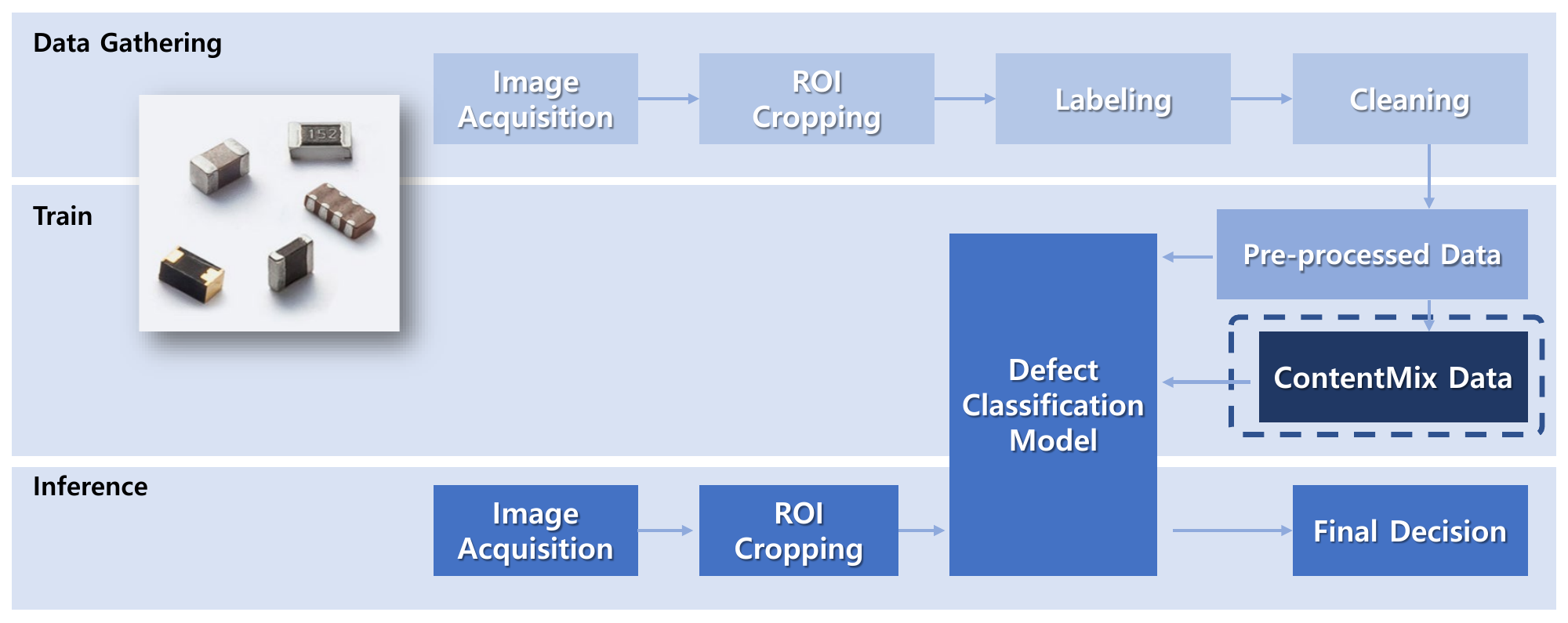}
    \end{tabular}}
    \caption{Overview of the component visual inspection system. The image dataset is intrinsically long-tailed data distribution. From this perspective, ContextMix generates new image data using existing labeled data in an inexpensive and efficient approach. The well-trained model by pre-processed and augmented data inference defective objects on each surface from acquired images in real-time.} 
    \label{fig:overviewVI}
\end{figure}
\subsection{Application to visual inspection}
To evaluate the effectiveness of our proposed method on industrial datasets, we demonstrate to apply the passive component defective inspection system and describe the overview of the system as shown in \cref{fig:overviewVI}.
In the data-gathering phase, the components are cuboid, and the inspection system acquires images per component surface. In general, the size of the raw image is larger than the size of the component. For the training efficiency of the classification network, acquired images are cropped, centering around the region of the inspection object, which is the component image. Then the cropped images are labeled to represent each class category by the experts of the productive process.
By the managers of the productive process, the data cleaning process is essential for the integrity of the labeling to prevent
mis-labeling since differences in work proficiency or a simple mistake.
As aforementioned, the gathered image datasets, named the pre-processed data, generally have long-tailed distributions that most image is labeled normal product since abnormal circumstance or defective products rarely happen. These imbalanced datasets, named pre-processed data, incur overfitted classification network, which is classified from most defective objects to normal products. In the training phase, we utilize basic data augmentation techniques using image transformations, such as random horizontal flipping, random vertical flipping, and weak color distortion. But random cropping, image translation, and image distortion are not adopted since the methods harm the characteristics of passive components.
However, the standard techniques are not sufficient to alleviate the overfitting issue, thus we use the pre-processed data to apply our proposed method to augment data. The network learns effective features to classify defective objects on the passive component, from pre-processed data and ContextMix-ed data. In the inference phase, the well-trained model classifies the acquired images from each of the different cameras on the inspection system. The images are cropped based on the region of interest, which is the inspection object. And the system should be synchronized with various hardware devices for the actual sorting system, physically. So we assign a simple module named final decision that merges each result of one product. And if one of the results is classified as defective, then it is a defective product. Then the module sends signals to the actual physical sorter whether the product is defective or not.

\section{Experimental Results} \label{sec:exps}
We evaluate ContextMix on various datasets with different network architectures on multiple tasks.
\cref{sec:exps_dataset} first introduces the datasets utilized to verify the performance of our proposed method. In \cref{sec:exps_classification}, we confirm the effectiveness of our proposed method on the image classification task with network architectures on the public benchmark datasets. Especially, \cref{sec:exps_classification_mlcc} presents the effectiveness of the industrial dataset from Samsung Electro-Mechanics (SEMCO). \cref{sec:exps_transfer} shows the compatibility of transfer learning through the object detection and segmentation tasks. We also show the weakly supervised object localization (WSOL) task in \cref{sec:exps_wsol}. Lastly, \cref{sec:exps_rbst_ece} provides robustness and calibration effectiveness of our proposed method.

\subsection{Dataset} \label{sec:exps_dataset}
We utilize various benchmark datasets to evaluate the performance of our proposed method. For image classification task, CIFAR-10/100~\cite{krizhevsky2009learning} and ImageNet~\cite{krizhevsky2012imagenet} datasets are utilized. CIFAR-10/100 datasets have 60,000 color images. 50,000 images are utilized for training, and 10,000 images are for validation. CIFAR-10 and CIFAR-100 have 10 classes and 100 classes, respectively. ImageNet dataset is a large-scale image dataset and has 1,000 classes. The dataset consists of 1.2 million natural images for training and 50,000 images for validation.
For object detection, we choose Pascal VOC dataset~\cite{everingham2010pascal, everingham2015pascal} and use 2007+2012 train-val data and 2007 test data for training and testing, respectively. To evaluate the performance of semantic segmentation, Pascal VOC 2012 dataset is utilized.
We experiment WSOL task and merge two benchmark datasets to utilize.
First, the dataset is CUB200-2011~\cite{wah2011caltech} which is a fine-grained image dataset and aims to classify 200 bird species. Another dataset is CUBv2~\footnote[1]{https://github.com/clovaai/wsolevaluation} that is released by~\cite{choe2020evaluating}. The dataset has collected 1,000 images from Flickr and has 1,000 annotations of bounding boxes of the images.
And we conduct verification of the robustness using the ImageNet validation dataset and ImageNet-A dataset~\cite{Hendrycks_2021_CVPR}. ImageNet-A consists of 200 classes and 7,500 natural adversarial samples for ImageNet-trained classifiers. The dataset, therefore, is designed for the ImageNet pre-trained model to have 0$\%$ accuracy.

\begin{figure}[t]
    \centering
    \resizebox{1.\linewidth}{!}{
    \setlength{\tabcolsep}{1pt}
    \begin{tabular}{cc}
        \includegraphics[width=0.95\linewidth]{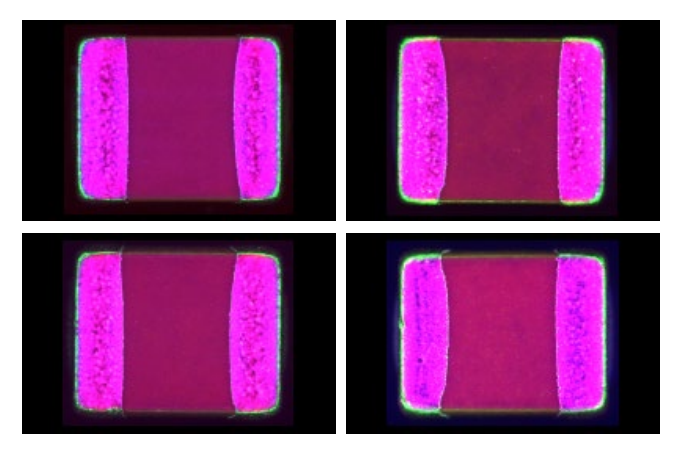} & \includegraphics[width=1.25\linewidth]{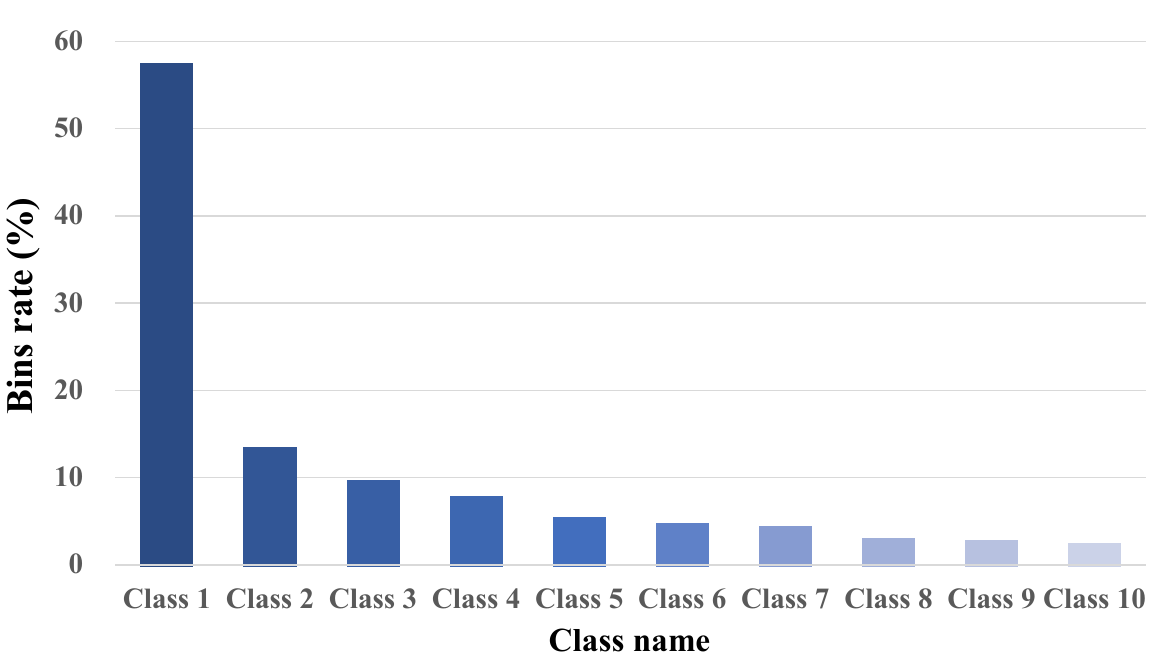}\\
        \fontsize{0.65cm}{0.65cm}\selectfont{(a) Normal Multi-layer ceramic capacitor (MLCC) images} &\fontsize{0.65cm}{0.65cm}\selectfont{(b) MLCC dataset distribution}\\
    \end{tabular}}
    \caption{Industrial dataset from Samsung Electro-Mechanics (SEMCO): (a) Normal MLCC images. (b) The dataset imbalanced distribution. Class 1 denotes normal MLCC products.}
    \label{fig:mlcc}
\end{figure}
\begin{table}[t]
\begin{center}
\footnotesize
\setlength{\tabcolsep}{1.0pt}
\begin{tabular}{p{0.2\linewidth}p{0.3\linewidth}P{0.12\linewidth}P{0.12\linewidth}P{0.12\linewidth}P{0.12\linewidth}}
    \toprule
    \multirow{2}{*}{Category} &\multirow{2}{*}{Dataset} &\multirow{2}{*}{Classes} &\multirow{2}{*}{Mean IR} &\multicolumn{2}{c}{Images}\\
    \cmidrule{5-6}
    & & & &Train &Valid\\
    \midrule
    \midrule
    \multirow{3}{*}{Benchmark} &CIFAR-10~\cite{krizhevsky2009learning} &10 &1.00 &50,000 &10,000\\
    &CIFAR-100~\cite{krizhevsky2009learning} &100 &1.00 &50,000 &10,000\\
    &ImageNet~\cite{krizhevsky2012imagenet} &1,000 &1.02 &1,281,167 &50,000\\
    \midrule
    Industrial &MLCC Inspection Image &10 &16.51 &52,304 &5,177\\
    \bottomrule
\end{tabular}
\end{center}
\caption{The datasets to evaluate classification tasks. Mean IR indicates averaged imbalance ratio, where 1 means perfectly balanced, and the imbalanced dataset records a value higher than 1. The Mean IR of industrial records at 16.51, is extremely imbalanced.}
\label{tab:dataset}
\end{table}

Moreover, we evaluate our proposed method on the real-world industrial dataset provided by SEMCO.
The notable distinction between benchmark datasets, like CIFAR and ImageNet, and industrial datasets, is found in the placement and size of the classified objects. In benchmark datasets, images are from different locations and diverse environments, encompassing variations in lighting and color tones, all while emphasizing the object of interest. As a result, the objects are vivid, well-focused, and usually situated at the center of the images, occupying a larger proportion of the frame compared to other areas containing unclassified objects or background elements. On the other hand, industrial datasets are gathered within controlled environments featuring steady lighting conditions, predetermined camera angles, and backgrounds with uniform tones. Additionally, the region of interest, the defect candidate region, is relatively smaller compared to the overall image size.

\cref{fig:mlcc}(a) shows multi-layer ceramic capacitors (MLCC) and the images are used in the inspection process to detect defects on the surface \cite{suh2017automatic}.
The dataset has ten classes, and class 1 represents the normal product.
The other class names are included sensitive information about SEMCO, so these are not public, instead are denoted from class 2 to class 10. And all the defective class names, categories, and data distributions are fictitious and not exact re-enactments.
However, we conduct that the dataset has a long-tailed distribution, as shown in \cref{fig:mlcc}(b). The well-balanced dataset records the mean imbalance ratio (Mean IR)~\cite{TAREKEGN2021107965} as 1. As for the poorly balanced dataset, the value is higher than 1. \cref{tab:dataset} presents that while the benchmark dataset is well-balanced, the Mean IR of the MLCC dataset is at 16.51 as extremely imbalanced.
\begin{table}[t]
\begin{center}
\footnotesize
\setlength{\tabcolsep}{1.0pt}
\begin{tabular}{p{0.13\linewidth}p{0.18\linewidth}p{0.3\linewidth}P{0.18\linewidth}P{0.18\linewidth}}
    \toprule
    Dataset &Model &Method &Top-1 Error~($\%$) &Top-5 Error~($\%$)\\
    \midrule
    \midrule
    \multirow{17}{*}{ImageNet} &\multirow{7}{*}{\shortstack{ResNet-50}} &Baseline$\ast$ &23.68 &7.05\\
    & &Cutout~\cite{devries2017improved}$\ast$ &22.93 &6.66\\
    & &Mixup~\cite{zhang2017mixup}$\ast$ &22.58 &6.40\\
    & &CutMix~\cite{yun2019cutmix}$\ast$ &21.40 &5.92\\
    & &PuzzleMix~\cite{kim2020puzzle}$\ast$ &21.24 &5.71\\
    & &SaliencyMix~\cite{uddin2021saliencymix}$\ast$ &21.26 &5.76\\
    & &\textbf{ContextMix} &\textbf{21.03} &\textbf{5.60}\\    
    \cmidrule{2-5}
    &\multirow{7}{*}{\shortstack{ResNet-101}} &Baseline$\ast$ &21.87 &6.29\\
    & &Cutout~\cite{devries2017improved}$\ast$ &20.72 &5.51\\
    & &Mixup~\cite{zhang2017mixup}$\ast$ &20.52 &5.28\\
    & &CutMix~\cite{yun2019cutmix}$\ast$ &20.17 &5.24\\
    & &PuzzleMix~\cite{kim2020puzzle} &19.91 &5.00\\
    & &SaliencyMix~\cite{uddin2021saliencymix}$\ast$ &20.09 &5.15\\
    & &\textbf{ContextMix} &\textbf{19.29} &\textbf{4.84}\\
    \cmidrule{2-5}
    &\multirow{3}{*}{\shortstack{ResNeXt-101\\(32$\times$8d)}} &Baseline$\ast$ &20.69 &5.47\\
    & &CutMix~\cite{yun2019cutmix} &18.72 &4.55\\
    & &\textbf{ContextMix} &\textbf{18.54} &\textbf{4.41}\\
    \bottomrule
\end{tabular}
\end{center}
\caption{ImageNet classification the best results based on ResNet-50, ResNet-101 and ResNeXt-101(32$\times$8d). $\ast$ denotes results reported in the CutMix, PuzzleMix, and SaliencyMix paper.}
\label{tab:image_classification_imagenet_cutmix}
\end{table}
\subsection{Benchmark Dataset Image Classification} \label{sec:exps_classification}
We first evaluate ContextMix performance on the ImageNet dataset for the image classification task. For a fair comparison, standard augmentation methods and hyperparameters are used the same as CutMix~\cite{yun2019cutmix} settings. We train all the models for 300 epochs with a batch size of 256 and an initial learning rate of 0.1. The learning rate is decayed by a factor of 0.1 at epochs 75, 150, and 225. The hyperparameter $\mathit{\alpha}$ for the beta distribution
is set to 1. We observe that ContextMix achieves the best result among the considered data augmentation methods for various network architectures, as shown in \cref{tab:image_classification_imagenet_cutmix}.
\begin{table}[!t]
\begin{center}
\footnotesize
\setlength{\tabcolsep}{1.0pt}
\begin{tabular}{p{0.12\linewidth}p{0.18\linewidth}p{0.35\linewidth}p{0.16\linewidth}p{0.16\linewidth}}
    \toprule
    Dataset& Model &Method &Top-1 Error~($\%$) &Top-5 Error~($\%$)\\
    \midrule
    \midrule
    \multirow{20}{*}[17pt]{CIFAR-100}&\multirow{8}{*}{\shortstack{PyramidNet-200\\($\alpha=240$)}} &Baseline$\ast$ &16.45 &3.69\\
    & &Cutout~\cite{devries2017improved}$\ast$ &16.53 &3.65\\
    & &Mixup~\cite{zhang2017mixup}$\ast$ &15.63 &3.99\\
    & &CutMix~\cite{yun2019cutmix}$\ast$ &14.47 &2.97\\
    & &PuzzleMix~\cite{kim2020puzzle} &{14.58$\pm$0.11} &2.93$\pm$0.13\\
    & &\textbf{ContextMix} &\textbf{14.13$\pm$0.20} &\textbf{2.83$\pm$0.10}\\
    \cmidrule{3-5}
    & &CutMix+ShakeDrop~\cite{yamada2019shakedrop} &12.94$\pm$0.08 &2.29$\pm$0.09\\
    & &\textbf{ContextMix+ShakeDrop}~\cite{yamada2019shakedrop} &\textbf{12.82$\pm$0.11} &\textbf{2.15$\pm$0.06}\\    
    \cmidrule{2-5}
    &\multirow{4}{*}{\shortstack{PyramidNet-110\\($\alpha=64$)}} &Baseline$\ast$ &19.85 &4.66\\
    & &CutMix~\cite{yun2019cutmix}$\ast$ &17.97 &3.83\\
    & &PuzzleMix~\cite{kim2020puzzle} &{18.37$\pm$0.18} &{3.67$\pm$0.22}\\
    & &\textbf{ContextMix} &\textbf{17.53$\pm$0.15} &\textbf{3.53$\pm$0.18}\\
    \cmidrule{2-5}
    &\multirow{4}{*}{PreActResnet-18} &Baseline$\ast$ &23.67 &8.98\\
    & &CutMix~\cite{yun2019cutmix}$\ast$ &23.20 &8.09\\
    & &PuzzleMix~\cite{kim2020puzzle}$\ast$ &19.62 &5.85\\
    & &\textbf{ContextMix} &\textbf{19.60$\pm$0.07} &\textbf{5.71$\pm$0.10}\\
    \cmidrule{2-5}
    &\multirow{4}{*}{\fontsize{0.28cm}{0.28cm}WRN28-10} &Baseline$\ast$ &21.14 &6.33\\
    & &CutMix\cite{yun2019cutmix}$\ast$ &17.50 &4.69\\
    & &PuzzleMix~\cite{kim2020puzzle}$\ast$ &15.95 &3.92\\
    & &\textbf{ContextMix} &\textbf{15.44$\pm$0.03} &\textbf{3.77$\pm$0.12}\\
    \midrule
    \multirow{6}{*}{\shortstack{CIFAR-10}} &\multirow{6}{*}{\shortstack{PyramidNet-200\\($\alpha=240$)}} &Baseline$\ast$ &3.85 &-\\
    & &Cutout~\cite{devries2017improved}$\ast$ &3.10 &-\\
    & &Mixup~\cite{zhang2017mixup}$\ast$ &3.09 &-\\
    & &CutMix~\cite{yun2019cutmix}$\ast$ &2.88 &0.11$\pm$0.03\\
    & &PuzzleMix~\cite{kim2020puzzle} &{2.90$\pm$0.04} &0.12$\pm$0.05\\
    & &\textbf{ContextMix} &\textbf{2.77$\pm$0.09} &\textbf{0.09$\pm$0.02}\\
    \bottomrule
\end{tabular}
\end{center}
\caption{CIFAR dataset classification results with various regularization methods. The results indicate mean and standard deviation over three runs. $\ast$ denotes averaged results reported in the CutMix, and PuzzleMix paper.}
\label{tab:image_classification_cifar100_cutmix}
\end{table}
On ResNet-50, our method records the best results as \textbf{21.03$\%$} Top-1 error and 5.60$\%$ top-5 error. ContextMix outperforms Cutout~\cite{devries2017improved}, Mixup~\cite{zhang2017mixup}, and CutMix by +1.90$\%$, +1.55$\%$ and +0.37$\%$, respectively. Our method stands over the methods such as PuzzleMix~\cite{kim2020puzzle} and SaliencyMix~\cite{uddin2021saliencymix} using saliency information by +0.21$\%$ and +0.23$\%$, respectively. In addition, ContextMix obtains the best results on ResNet-101 as \textbf{19.29$\%$}. In particular, compared with PuzzleMix and SaliencyMix, we note improvements of \textbf{+0.62$\%$} and \textbf{+0.80$\%$} without additional inference. Also, on a deeper architecture, ResNeXt-101(32$\times$8d)~\cite{xie2017aggregated}, ContextMix outperforms CutMix by \textbf{+0.18$\%$}.

\begin{table}[t]
\begin{center}
\footnotesize
\setlength{\tabcolsep}{1.0pt}
\begin{tabular}{p{0.1\linewidth}p{0.13\linewidth}p{0.16\linewidth}P{0.18\linewidth}P{0.18\linewidth}P{0.22\linewidth}}
    \toprule
    Dataset &Model &Method &Top-1 Error~($\%$) &Top-5 Error~($\%$) &Macro F1 Score~($\%$)\\    
    \midrule
    \midrule
    \multirow{7}{*}[5pt]{\shortstack{MLCC\\dataset}} &\multirow{7}{*}[5pt]{ResNet-18} &Baseline &6.52$\pm$0.66 &0.69$\pm$0.11 &0.912$\pm$0.009\\
    & &Cutout~\cite{devries2017improved} &7.71$\pm$0.63 &0.59$\pm$0.06 &0.927$\pm$0.005\\
    & &Mixup~\cite{zhang2017mixup} &\textbf{5.79$\pm$0.31} &0.54$\pm$0.10 &0.939$\pm$0.003\\
    & &CutMix~\cite{yun2019cutmix} &7.81$\pm$0.40 &0.60$\pm$0.07 &0.901$\pm$0.004\\
    & &PuzzleMix~\cite{kim2020puzzle} &6.81$\pm$0.32 &0.58$\pm$0.10 &0.927$\pm$0.011\\
    & &SaliencyMix~\cite{uddin2021saliencymix} &7.91$\pm$0.19 &0.54$\pm$0.21 &0.920$\pm$0.005\\
    & &\textbf{ContextMix} &5.80$\pm$0.27 &\textbf{0.51$\pm$0.09} &\textbf{0.948$\pm$0.004}\\
    \bottomrule
\end{tabular}
\end{center}
\caption{MLCC dataset classification results with various data augmentation methods using ResNet-18. The results indicate mean and standard deviation over three runs.}
\label{tab:image_classification_mlcc}
\end{table}
On CIFAR datasets, we follow the augmentation procedures and hyperparameter settings in CutMix and PuzzleMix for fair comparisons, respectively.
We train for 300 epochs with a batch size of 64 on PyramidNet-200 and PyramidNet-110, with a widening factor $\alpha$ of 240 and 64, respectively.
The initial learning rate is set to 0.25 and decayed by 0.1 at 150 and 225 epochs.
For PreActResNet-18~\cite{he2016identity} and WRN28-10~\cite{BMVC2016_87}, we set hyperparameters following PuzzleMix. ContextMix is trained using PreActResNet-18 for 400 epochs and WRN28-10 for 1,200 epochs. The initial learning rate is set to 0.1 and decayed by 0.1 at 200 and 300 epochs for PreActResNet-18 and 400 and 800 for WRN28-10.
As presented in \cref{tab:image_classification_cifar100_cutmix}, ContextMix shows also improved performances on CIFAR datasets. All the experiments on CIFAR datasets are evaluated three times.
On the CIFAR-100,
ContextMix obtains \textbf{14.13$\%$} Top-1 error. We also evaluate ContextMix along with ShakeDrop~\cite{yamada2019shakedrop} applied during training and obtain 12.82$\%$ Top-1 error. In addition, we explore PyramidNet-110~\cite{han2017deep} and enhance Top-1 accuracy by +0.44$\%$. 
Also, our method shows 2$\%$  and 5$\%$  improvement on the PreActResNet-18 and WRN28-10, respectively.
Also, our method shows improvement as compared to other regularization methods on CIFAR-10.
\subsection{Industrial dataset Classification} \label{sec:exps_classification_mlcc}
We evaluate the impact of our proposed method on the industrial, which is the MLCC dataset generated by SEMCO.
The dataset is trained during 150 epochs with a batch size of 64. The initial learning rate is 0.1, then decayed by 10 every 30 epochs.
\cref{tab:image_classification_mlcc} presents the results of ContextMix and other compared methods. We measure the performances using metrics which are the Top-1 error rate and the macro F1 Score. The F1 score is calculated using recall and precision and is defined as ${2PR}/{(P+R)}$, where P and R indicate precision and recall, respectively. The macro F1 score, a widely utilized metric, accounts for both majority and minority class performance relative to each other, proving particularly effective in measuring model accuracy on imbalanced datasets. For that reasons, we evaluated the performance of our proposed method not only using top-1 error but also macro F1 score on the real industrial dataset MLCC.

ContextMix records the second-best performance next to Mixup.
The gap in the Top-1 error rates between ContextMix and Mixup is 0.01$\%$ only.
Except for Mixup, other compared methods harm performance and record under-performance over baseline.
We hypothesize that the defect object size is enough smaller than the whole size of MLCC.
It means that Mixup and our proposed method preserve important defect features and represent mixed new image data better than other compared methods.
Specifically, Cutout is possible to lose the defect region by random cropping. CutMix also inherits the chance by random cropping and pasting. The region that cropping and pasting of PuzzleMix and SaliencyMix are acquired on saliency information. Getting incorrect saliency information leads to losing crucial defect regions by cropping and pasting.
From this perspective, the baseline, which does not employ data augmentation approaches, could not miss the defective regions fundamentally, resulting in better performance than Cutout, CutMix, and the saliency-based model.
However, Mixup can preserve the features by exploiting overlapped two different images by mixing ratio. And ContextMix may eliminate a defective region through cropping but retain other defect features by pasting from various resized images.
In \cref{sec:abl_saliency} and \cref{sec:abl_sizewise}, we elaborate on the drawbacks of saliency-based methods and provide a detailed account of the benefits they offer in terms of classification accuracy through the utilization of resized images. Moreover, considering the dataset has long-tailed distribution, the macro F1 score is a better metric than the Top-1 error rate. ContextMix records the best F1 score than other considered methods. 
\begin{table}[!t]
\footnotesize
\begin{center}
\setlength{\tabcolsep}{1pt}
\renewcommand{\arraystretch}{1.0}
\begin{tabular}{p{0.33\linewidth}P{0.21\linewidth}P{0.21\linewidth}P{0.01\linewidth}P{0.21\linewidth}}
    \toprule
    \multirow{2}{*}[-1pt]{~Method} &\multicolumn{2}{c}{Detection(mAP)} & &Segmentation(mIOU)\\
    \cmidrule{2-3}\cmidrule{5-5}
    &Faster R-CNN~\cite{ren2015faster}&SSD~\cite{liu2016ssd}& &DeepLabv3~\cite{Chen_2018_ECCV}\\
    \midrule
    \midrule
    Baseline &80.87 &76.13 & &75.80\\
    Cutout~\cite{devries2017improved} &80.44 &76.35 & &76.56\\
    Mixup~\cite{zhang2017mixup} &80.82 &76.79 & &77.04\\
    CutMix~\cite{yun2019cutmix} &81.04 &77.17 & &77.31\\
    PuzzleMix~\cite{kim2020puzzle} &78.20 &76.10 & &77.29\\
    SaliencyMix~\cite{uddin2021saliencymix} &79.87 &77.18 & &76.86\\
    \textbf{ContextMix} &\textbf{81.49} &\textbf{77.75} & &\textbf{78.10}\\
    \bottomrule
\end{tabular}
\end{center}
\caption{Detection and Segmentation task results with ResNet-50 trained on Pascal VOC 2007 and 2012.}
\label{tab:detection}
\end{table}

\subsection{Transfer Learning of Pre-trained Model} \label{sec:exps_transfer}
ImageNet pre-trained models are widely used as standard initialization for many computer vision applications. To demonstrate the generalization ability of ContextMix-trained network, we evaluate the model on object detection task and semantic segmentation task. The results show that ContextMix-trained feature extractors have higher quality and thus can be effectively transferred to other vision tasks.

Following the evaluation procedure of Pascal VOC 2007, we use mAP@0.5 as the evaluation metric.
We use different detection frameworks:
Faster R-CNN~\cite{ren2015faster}\footnote[2]{https://github.com/open-mmlab/mmdetection} with feature pyramid network (FPN)~\cite{lin2017feature} and single shot multibox detector (SSD) \cite{liu2016ssd}\footnote[3]{https://github.com/yqyao/SSD$\_$Pytorch}. For the framework, we use the ResNet-50 backbone and compare the detection performances of the network initialized with different ImageNet-trained weights. \cref{tab:detection} presents that the ContextMix-trained weights give more performance boost compared to other data augmentation methods.

Semantic segmentation is a task that requires a network to predict correct semantic labels for every pixel of an image. We use Pascal VOC 2012 dataset for training and validation and the architecture proposed in DeepLabv3~\cite{Chen_2018_ECCV}\footnote[4]{https://github.com/jfzhang95/pytorch-deeplab-xception}. \cref{tab:detection} presents that ContextMix-trained method weights produce the best performance among the competitive methods.

Noteworthy, unlike the results of the classification task, the performances of the baseline in the detection and segmentation tasks are recorded competitively with the saliency information-based approach. This indicates that the methods are influenced by the characteristics of the datasets.
\begin{table}[t]
\footnotesize
\begin{center}
\setlength{\tabcolsep}{1pt}
\renewcommand{\arraystretch}{1.0}
\begin{tabular}{p{0.25\linewidth}P{0.24\linewidth}P{0.21\linewidth}P{0.25\linewidth}}
    \toprule
    \multirow{2}{*}[-2pt]{Method} &\multirow{2}{*}[-2pt]{Dataset} &\multicolumn{2}{c}{Localization}\\
    \cmidrule{3-4}
    & &Top-1 Accuracy~($\%$) &MaxBoxAcc~($\%$)~\cite{choe2020evaluating}\\
    \midrule
    \midrule
    Mixup~\cite{zhang2017mixup} &\multirow{4}{*}[-2pt]{\shortstack{CUB200-2011\\+\\CUBv2}} &52.78 &69.38\\
    CutMix~\cite{yun2019cutmix} & &\textbf{56.40} &73.34\\
    PuzzleMix~\cite{kim2020puzzle} & &54.74 &69.81\\
    \textbf{ContextMix} & &55.47 &\textbf{73.80}\\
    \bottomrule
\end{tabular}
\end{center}
\caption{Weakly Supervised Object Localization result and with ResNet-50. The results indicate an average value over three runs.}
\label{tab:wsol}
\end{table}
\begin{figure*}[!t]
    \centering
    \small
    \resizebox{1.\linewidth}{!}{
    \rotatebox{90}{
    \setlength{\tabcolsep}{1.0pt}
    \begin{tabular}{||cccccccc||}
        \fontsize{15cm}{15cm}\selectfont\textbf{ContextMix}
        &\fontsize{15cm}{15cm}\selectfont\textbf{PuzzleMix}
        &\fontsize{15cm}{15cm}\selectfont\textbf{CutMix}
        &\fontsize{15cm}{15cm}\selectfont\textbf{Clean}
        &\fontsize{15cm}{15cm}\selectfont\textbf{ContextMix}
        &\fontsize{15cm}{15cm}\selectfont\textbf{PuzzleMix}
        &\fontsize{15cm}{15cm}\selectfont\textbf{CutMix}
        &\fontsize{15cm}{15cm}\selectfont\textbf{Clean}\\

        \rotatebox{270}{\includegraphics[width=0.6\linewidth]{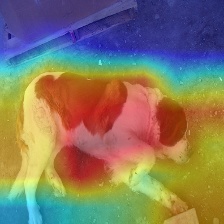}}
        &\rotatebox{270}{\includegraphics[width=0.6\linewidth]{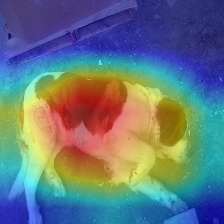}}
        &\rotatebox{270}{\includegraphics[width=0.6\linewidth]{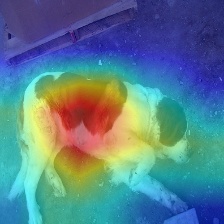}}
        &\rotatebox{270}{\includegraphics[width=0.6\linewidth]{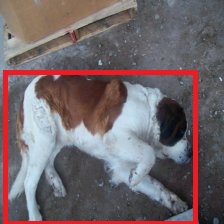}}
        &\rotatebox{270}{\includegraphics[width=0.6\linewidth]{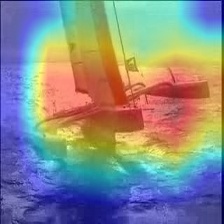}}
        &\rotatebox{270}{\includegraphics[width=0.6\linewidth]{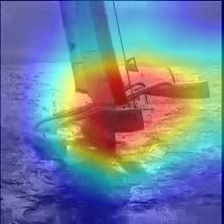}}
        &\rotatebox{270}{\includegraphics[width=0.6\linewidth]{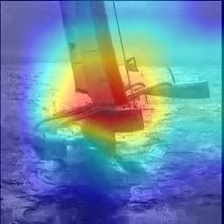}}
        &\rotatebox{270}{\includegraphics[width=0.6\linewidth]{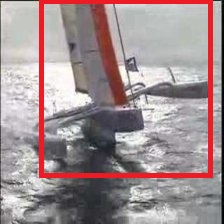}}\\
        
        \rotatebox{270}{\includegraphics[width=0.6\linewidth]{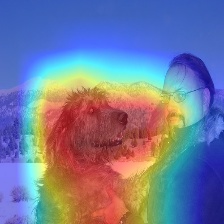}}
        &\rotatebox{270}{\includegraphics[width=0.6\linewidth]{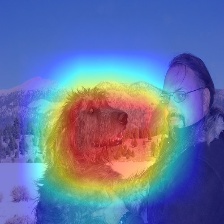}}
        &\rotatebox{270}{\includegraphics[width=0.6\linewidth]{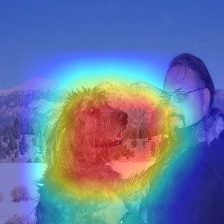}}
        &\rotatebox{270}{\includegraphics[width=0.6\linewidth]{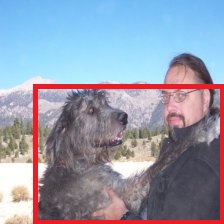}}
        &\rotatebox{270}{\includegraphics[width=0.6\linewidth]{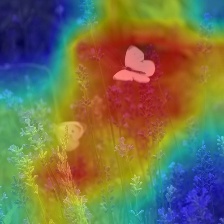}}
        &\rotatebox{270}{\includegraphics[width=0.6\linewidth]{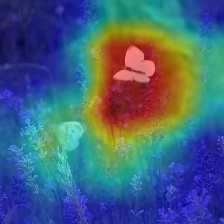}}
        &\rotatebox{270}{\includegraphics[width=0.6\linewidth]{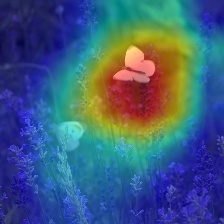}}
        &\rotatebox{270}{\includegraphics[width=0.6\linewidth]{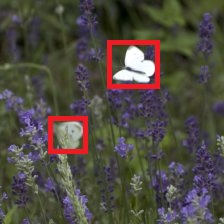}}\\
        
        \rotatebox{270}{\includegraphics[width=0.6\linewidth]{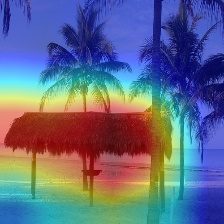}}
        &\rotatebox{270}{\includegraphics[width=0.6\linewidth]{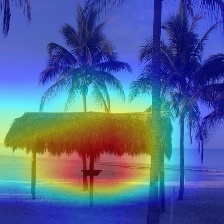}}
        &\rotatebox{270}{\includegraphics[width=0.6\linewidth]{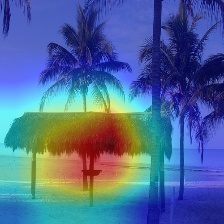}}
        &\rotatebox{270}{\includegraphics[width=0.6\linewidth]{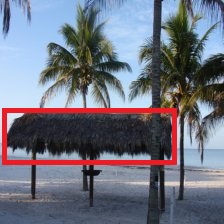}}        
        &\rotatebox{270}{\includegraphics[width=0.6\linewidth]{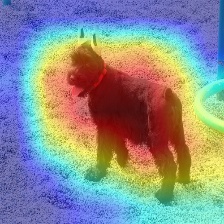}}
        &\rotatebox{270}{\includegraphics[width=0.6\linewidth]{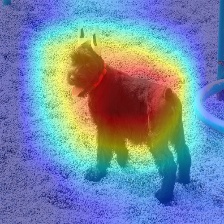}}
        &\rotatebox{270}{\includegraphics[width=0.6\linewidth]{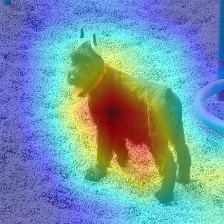}}
        &\rotatebox{270}{\includegraphics[width=0.6\linewidth]{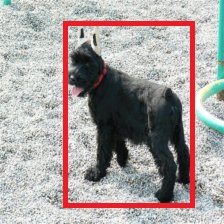}}\\
        
        \rotatebox{270}{\includegraphics[width=0.6\linewidth]{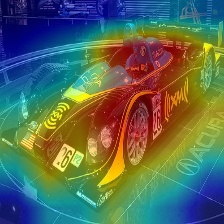}}
        &\rotatebox{270}{\includegraphics[width=0.6\linewidth]{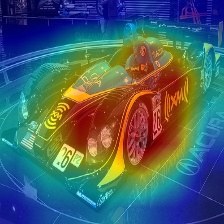}}
        &\rotatebox{270}{\includegraphics[width=0.6\linewidth]{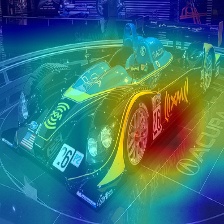}}
        &\rotatebox{270}{\includegraphics[width=0.6\linewidth]{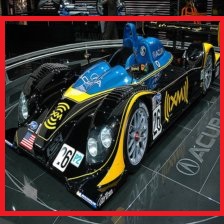}}
        &\rotatebox{270}{\includegraphics[width=0.6\linewidth]{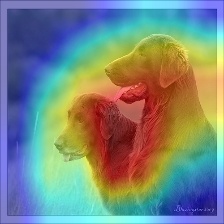}}
        &\rotatebox{270}{\includegraphics[width=0.6\linewidth]{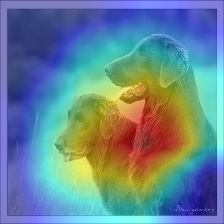}}
        &\rotatebox{270}{\includegraphics[width=0.6\linewidth]{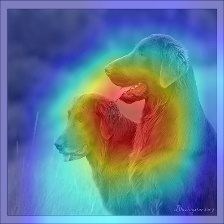}}
        &\rotatebox{270}{\includegraphics[width=0.6\linewidth]{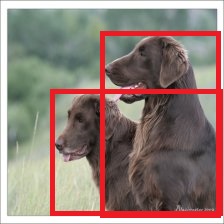}}\\
        
        \rotatebox{270}{\includegraphics[width=0.6\linewidth]{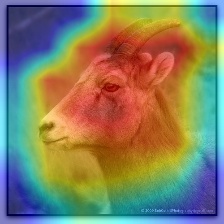}}
        &\rotatebox{270}{\includegraphics[width=0.6\linewidth]{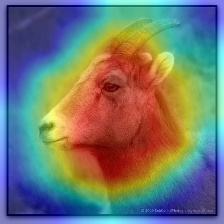}}
        &\rotatebox{270}{\includegraphics[width=0.6\linewidth]{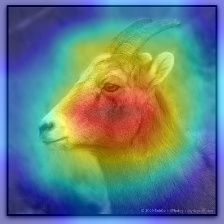}}
        &\rotatebox{270}{\includegraphics[width=0.6\linewidth]{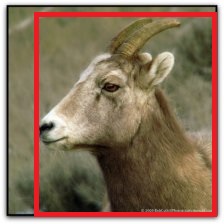}}
        &\rotatebox{270}{\includegraphics[width=0.6\linewidth]{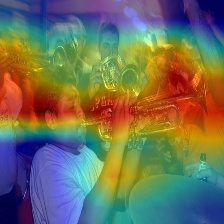}}
        &\rotatebox{270}{\includegraphics[width=0.6\linewidth]{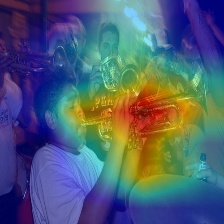}}
        &\rotatebox{270}{\includegraphics[width=0.6\linewidth]{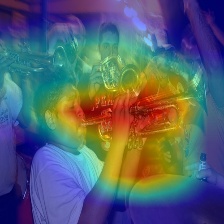}}
        &\rotatebox{270}{\includegraphics[width=0.6\linewidth]{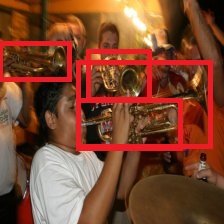}}\\
        
        \rotatebox{270}{\includegraphics[width=0.6\linewidth]{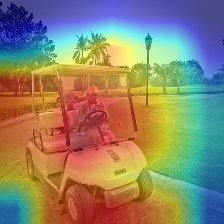}}
        &\rotatebox{270}{\includegraphics[width=0.6\linewidth]{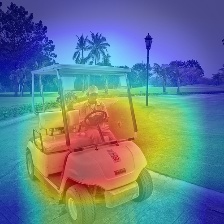}}
        &\rotatebox{270}{\includegraphics[width=0.6\linewidth]{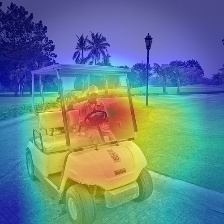}}
        &\rotatebox{270}{\includegraphics[width=0.6\linewidth]{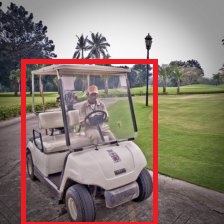}}
        &\rotatebox{270}{\includegraphics[width=0.6\linewidth]{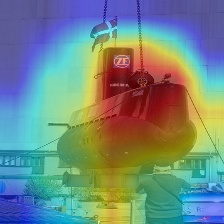}}
        &\rotatebox{270}{\includegraphics[width=0.6\linewidth]{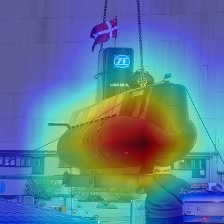}}
        &\rotatebox{270}{\includegraphics[width=0.6\linewidth]{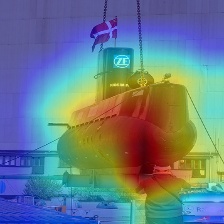}}
        &\rotatebox{270}{\includegraphics[width=0.6\linewidth]{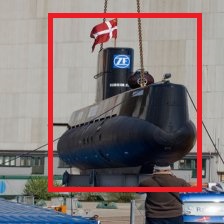}}\\
        
        \rotatebox{270}{\includegraphics[width=0.6\linewidth]{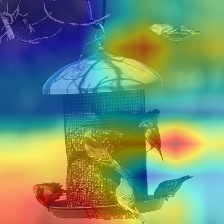}}
        &\rotatebox{270}{\includegraphics[width=0.6\linewidth]{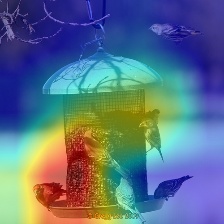}}
        &\rotatebox{270}{\includegraphics[width=0.6\linewidth]{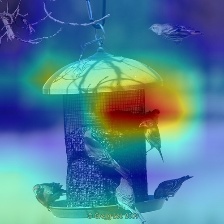}}
        &\rotatebox{270}{\includegraphics[width=0.6\linewidth]{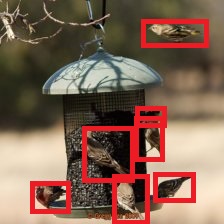}}
        &\rotatebox{270}{\includegraphics[width=0.6\linewidth]{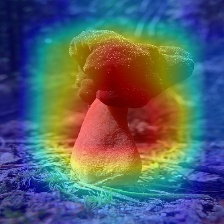}}
        &\rotatebox{270}{\includegraphics[width=0.6\linewidth]{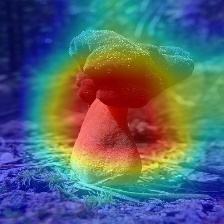}}
        &\rotatebox{270}{\includegraphics[width=0.6\linewidth]{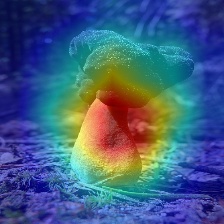}}
        &\rotatebox{270}{\includegraphics[width=0.6\linewidth]{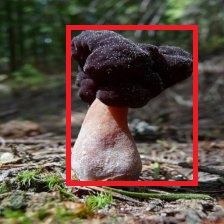}}\\
        
        \rotatebox{270}{\includegraphics[width=0.6\linewidth]{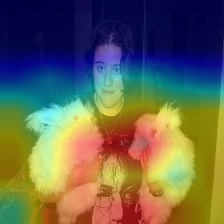}}
        &\rotatebox{270}{\includegraphics[width=0.6\linewidth]{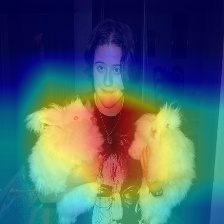}}
        &\rotatebox{270}{\includegraphics[width=0.6\linewidth]{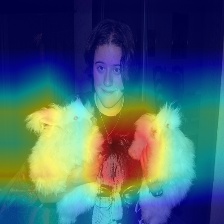}}
        &\rotatebox{270}{\includegraphics[width=0.6\linewidth]{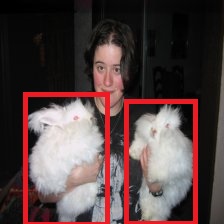}}
        &\rotatebox{270}{\includegraphics[width=0.6\linewidth]{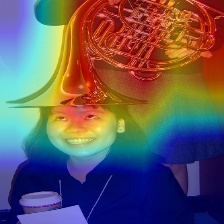}}
        &\rotatebox{270}{\includegraphics[width=0.6\linewidth]{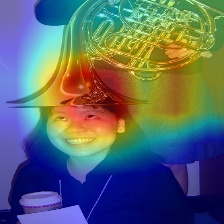}}
        &\rotatebox{270}{\includegraphics[width=0.6\linewidth]{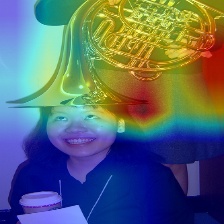}}
        &\rotatebox{270}{\includegraphics[width=0.6\linewidth]{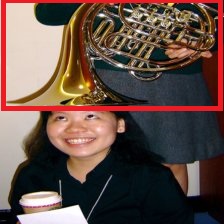}}\\
    \end{tabular}}}
    \caption{The Class activation map (CAM)~\cite{zhou2016learning} results of ContextMix present that covers the objects more precise and broader with high scores(reddish region) than CutMix and PuzzleMix. Red boxes indicate ground truth object location.}
    \label{fig:cam_cutmix_ours}
\end{figure*}
\subsection{Weakly Supervised Object Localization (WSOL)} \label{sec:exps_wsol}
For WSOL evaluation, the metrics are Top-1 accuracy and MaxBoxAcc~\cite{choe2020evaluating}. Top-1 accuracy measures IOU@0.5 between the predicted class bounding box and ground truth location. MaxBoxAcc measures the maximal box accuracy at the optimal threshold of heatmap with IOU@0.5. As shown in Table~\ref{tab:wsol}, the Top-1 localization accuracy of ContextMix under-performs than CutMix. Still, it shows improved results than PuzzleMix, and ContextMix records the better performance of MaxBoxAcc than CutMix and PuzzleMix.
\begin{table}[!t]
\footnotesize
\begin{center}
\setlength{\tabcolsep}{1pt}
\renewcommand{\arraystretch}{1.0}
\begin{tabular}{p{0.35\linewidth}P{0.2\linewidth}P{0.2\linewidth}P{0.01\linewidth}P{0.2\linewidth}}
    \toprule
    \multirow{3}{*}[-2pt]{Method} &\multicolumn{2}{c}{Robustness} & &Calibration\\
    \cmidrule{2-3}\cmidrule{5-5}
    &\multicolumn{2}{c}{Top-1 Accuracy~($\%$)} & &\multirow{2}{*}[-2pt]{ECE($\%$)}\\
    \cmidrule{2-3}
    &FGSM &ImageNet-A & &\\
    \midrule
    \midrule
    Baseline &8.2 &0.00 & &3.7\\
    Cutout~\cite{devries2017improved} &11.5 &4.24 & &5.4\\
    Mixup~\cite{zhang2017mixup} &24.4 &5.68 & &4.0\\
    CutMix~\cite{yun2019cutmix} &33.7 &7.25 & &1.8\\
    PuzzleMix~\cite{kim2020puzzle} &28.5 &6.80 & &4.1\\
    SaliencyMix~\cite{uddin2021saliencymix} &34.6 &8.00 & &\textbf{1.6}\\
    \textbf{ContextMix} &\textbf{37.3} &\textbf{9.63} & &2.8\\
    \bottomrule
\end{tabular}
\end{center}
\caption{Robustness by FGSM and ImageNet-A, and Calibration results with ResNet-50.}
\label{tab:adversarial_CutMix}
\end{table}

We claim that this is a clue that ContextMix learned the object and contextual information.
\cref{fig:cam_cutmix_ours} shows the results of CAM~\cite{zhou2016learning} by CutMix-, PuzzleMix-, and ContextMix-trained networks on the ImageNet validation dataset. More specifically, from the perspective of CAM's activated regions, CutMix focuses on specific areas of the object, and PuzzleMix shows expanded areas of activated than CutMix but shrank areas than our method.
And ContextMix covers the whole region of the object, including partial background, and the regions are the most expansive than other compared methods.

\subsection{Robustness and Calibration} \label{sec:exps_rbst_ece}
We experiment to ensure that our proposed method is sufficiently robust against adversarial attacks on ImageNet validation dataset on the well-known fast gradient sign method (FGSM)~\cite{goodfellow2014explaining}\footnote[5]{https://github.com/Harry24k/adversarial-attacks-pytorch}.
Using ImageNet-A dataset, we evaluate robustness from natural examples. Calibrated probability estimates are used for empirical risk analysis, which prevents overly confident predictions. We, also, evaluate the risk using expected calibration error (ECE)~\cite{guo2017calibration}.

As shown in \cref{tab:adversarial_CutMix}, the robustness of FGSM and ImageNet-A presents that ContextMix shows improved accuracy over the considered competitive methods. Also, even though PuzzleMix invests additional computational costs to acquire saliency information, it shows a higher over-confidence value than the Mixup method. CutMix-trained and SaliencyMix-trained network performs under-confident than ContextMix.
However, we evaluate that ContextMix shows enough under-confidence than other considered methods, relatively.

\section{Discussion and Ablation Study} \label{sec:discussion}
In this section, \cref{sec:abl_saliency} question generated data by saliency-based methods. The methods incur additional computational costs for acquiring the salient regions, whereas the regions are not always pointed to objects. Then we verify the key factors of ContextMix by evaluating various ablation studies. In \cref{sec:abl_sizewise}, we analyze the performances of objects size-wise on classification and object detection tasks. Through \cref{sec:abl_rbst_blur}, we confirm the resized images preserve enough information to recognize objects. \cref{sec:abl_resizeratio} measure influences of resizing ratio. \cref{sec:abl_variant} introduce our proposed method variants and evaluate. At Last, we confirm the suitability of randomly selected pasting regions in \cref{sec:abl_fixed}.
\begin{figure}[!ht]
    \centering
    \resizebox{0.95\linewidth}{!}{
    \setlength{\tabcolsep}{1pt}
    \begin{tabular}{cccccccc}
        \includegraphics[width=0.5\linewidth]{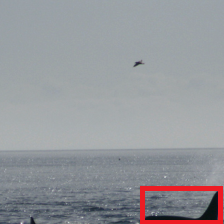} 
        &\includegraphics[width=0.5\linewidth]{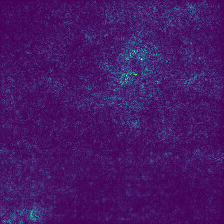} 
        &\multirow{2}{*}[80pt]{\includegraphics[width=0.5\linewidth]{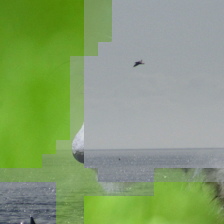}} 
        &\includegraphics[width=0.5\linewidth]{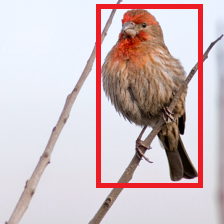} 
        &\includegraphics[width=0.5\linewidth]{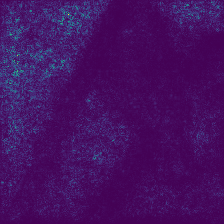} 
        &\multirow{2}{*}[80pt]{\includegraphics[width=0.5\linewidth]{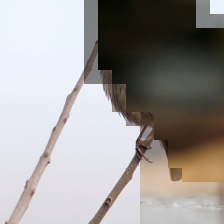}}\\
        \includegraphics[width=0.5\linewidth]{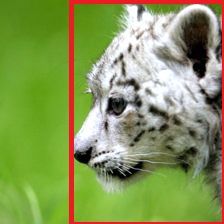} 
        &\includegraphics[width=0.5\linewidth]{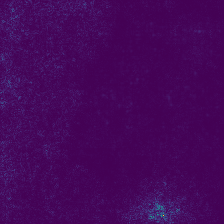} & 
        &\includegraphics[width=0.5\linewidth]{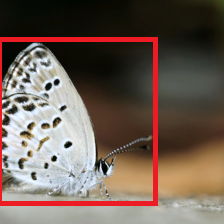} 
        &\includegraphics[width=0.5\linewidth]{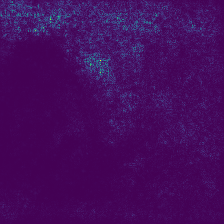} &\\
        
        \includegraphics[width=0.5\linewidth]{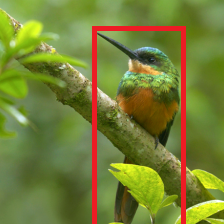} 
        &\includegraphics[width=0.5\linewidth]{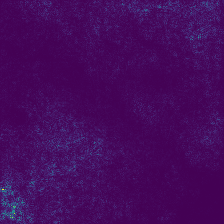} 
        &\multirow{2}{*}[80pt]{\includegraphics[width=0.5\linewidth]{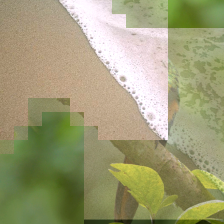}} 
        &\includegraphics[width=0.5\linewidth]{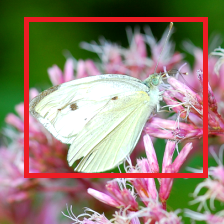} 
        &\includegraphics[width=0.5\linewidth]{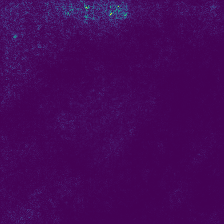} 
        &\multirow{2}{*}[80pt]{\includegraphics[width=0.5\linewidth]{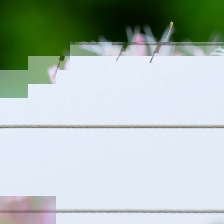}}\\
        \includegraphics[width=0.5\linewidth]{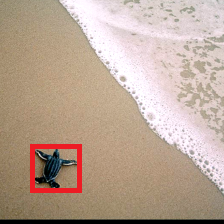} 
        &\includegraphics[width=0.5\linewidth]{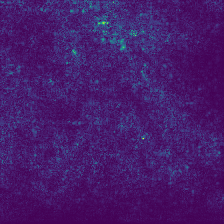} & 
        &\includegraphics[width=0.5\linewidth]{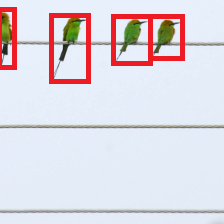} 
        &\includegraphics[width=0.5\linewidth]{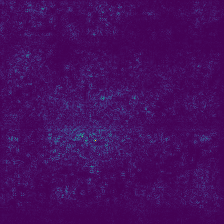} &\\

        \includegraphics[width=0.5\linewidth]{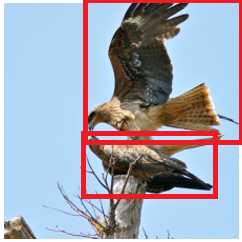} 
        &\includegraphics[width=0.5\linewidth]{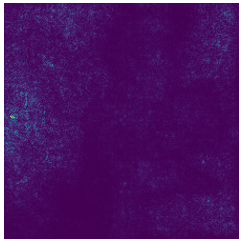} 
        &\multirow{2}{*}[80pt]{\includegraphics[width=0.5\linewidth]{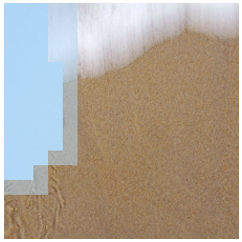}}
        &\includegraphics[width=0.5\linewidth]{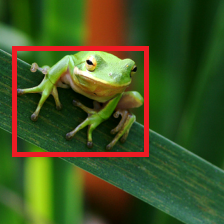} 
        &\includegraphics[width=0.5\linewidth]{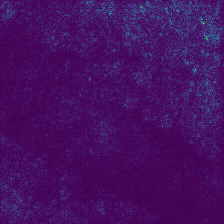} 
        &\multirow{2}{*}[80pt]{\includegraphics[width=0.5\linewidth]{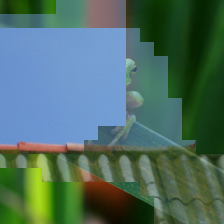}}\\
        \includegraphics[width=0.5\linewidth]{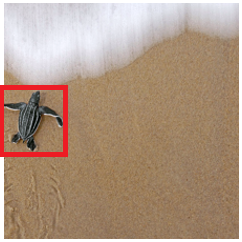} 
        &\includegraphics[width=0.5\linewidth]{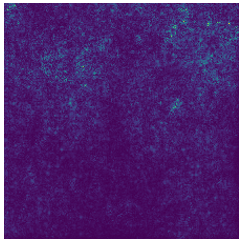} &
        &\includegraphics[width=0.5\linewidth]{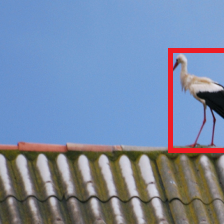} 
        &\includegraphics[width=0.5\linewidth]{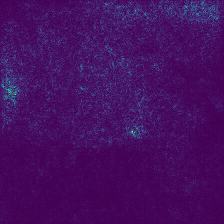} &\\
        
        \includegraphics[width=0.5\linewidth]{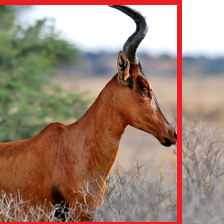}
        &\includegraphics[width=0.5\linewidth]{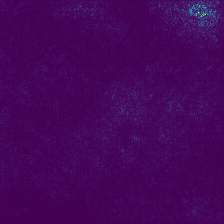} 
        &\multirow{2}{*}[80pt]{\includegraphics[width=0.5\linewidth]{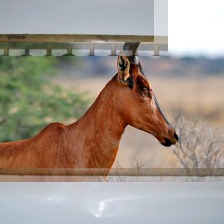}}
        &\includegraphics[width=0.5\linewidth]{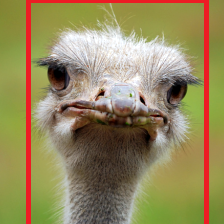} 
        &\includegraphics[width=0.5\linewidth]{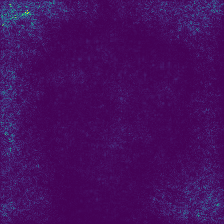} 
        &\multirow{2}{*}[80pt]{\includegraphics[width=0.5\linewidth]{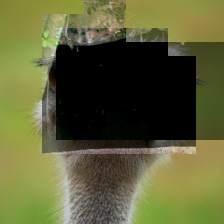}}\\
        \includegraphics[width=0.5\linewidth]{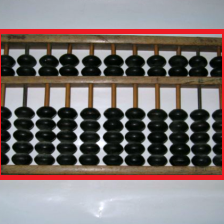} 
        &\includegraphics[width=0.5\linewidth]{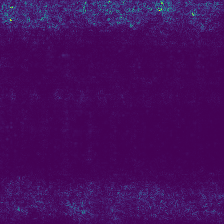} &
        &\includegraphics[width=0.5\linewidth]{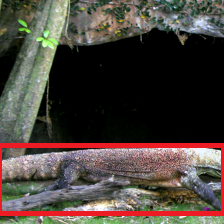} 
        &\includegraphics[width=0.5\linewidth]{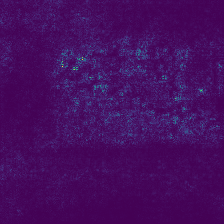} &\\        

        \fontsize{1.8cm}{1.8cm}\selectfont{Input Image}
        &\fontsize{1.8cm}{1.8cm}\selectfont{Saliency Map}
        &\fontsize{1.8cm}{1.8cm}\selectfont{Mixed Image}
        &\fontsize{1.8cm}{1.8cm}\selectfont{Input Image}
        &\fontsize{1.8cm}{1.8cm}\selectfont{Saliency Map}
        &\fontsize{1.8cm}{1.8cm}\selectfont{Mixed Image}\\
    \end{tabular}}
    \caption{Failure cases by the Saliency-based Method with ResNet-50 on ImageNet dataset.}
    \label{fig:saliency_method}
\end{figure}

\subsection{Drawback of the Saliency-based Method} \label{sec:abl_saliency}
Saliency information~\cite{wang2015deep, simonyan2013deep} and CAM~\cite{zhou2016learning} method help explain the region where the network is concentrated on recognizing objects in images. The premise of saliency-based data augmentation methods is the assumption that salient regions are objects. In most cases, the salient region highlights the area of the actual objects.

However, the problem is that the highlighted areas are on the background area, not objects. \cref{fig:saliency_method} presents these worst examples. Each object in the input images is shown clearly.
Each object in the input images is confirmed clearly. While the highlighted regions by the PuzzleMix-trained network~\cite{kim2020puzzle}\footnote[6]{https://github.com/snu-mllab/PuzzleMix/tree/master/imagenet} are not located in the region of the object. As a result, generated new images by saliency methods contain only the background from each clean image.

On the other hand, ContextMix-ed images have no chance to mix only background images; rather always have object information from the resized whole image and avoid the drawback. Furthermore, generated images by saliency-based methods can contain only one object information which means the image is mixed with objects and background from different images, respectively. This case also contradicts that the salient regions should always be object regions by the methods scheme.
\begin{figure}[t]
    \centering
    \resizebox{0.8\linewidth}{!}{
    \setlength{\tabcolsep}{1pt}
    \begin{tabular}{c}
        \includegraphics[width=1\linewidth]{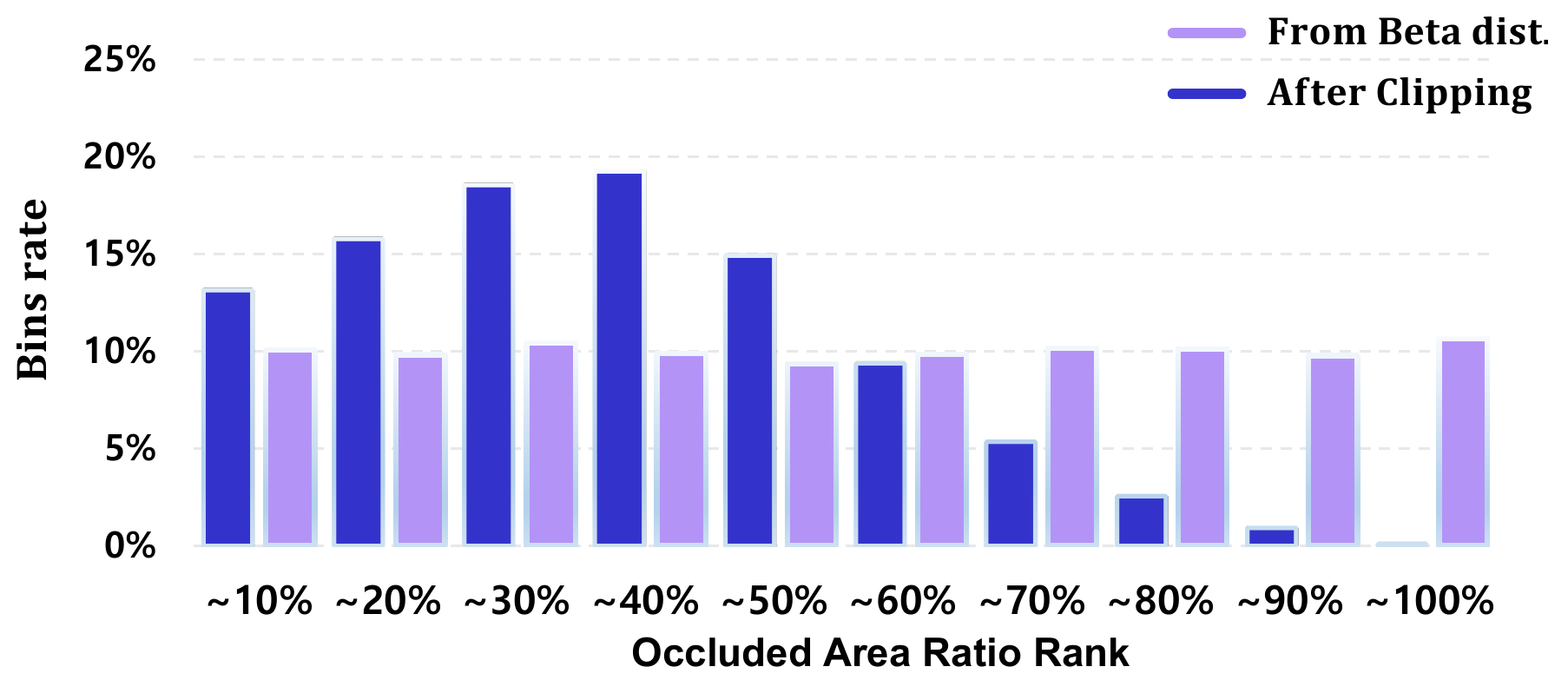}\\
    \end{tabular}}
    \caption{Occluded areas Histogram of initial values from Beta Distribution and after Clipping operation for fitting to image size.}
    \label{fig:classification_size}
\end{figure}
\begin{table}[!ht]
\footnotesize
\begin{center}
\setlength{\tabcolsep}{1pt}
\renewcommand{\arraystretch}{0.95}
\begin{tabular}{p{0.2\linewidth}P{0.11\linewidth}P{0.08\linewidth}P{0.08\linewidth}P{0.08\linewidth}P{0.01\linewidth}P{0.11\linewidth}P{0.08\linewidth}P{0.08\linewidth}P{0.08\linewidth}}
    \toprule
    &\multicolumn{4}{c}{Classification} & &\multicolumn{4}{c}{Detection}\\
    \cmidrule{2-5}\cmidrule{7-10}
    \multirow{2}{*}[-2pt]{Method} & &\multicolumn{3}{c}{Top-1 Error~($\%$)} & & &\multicolumn{3}{c}{mAP}\\
    \cmidrule{3-5}\cmidrule{8-10}
    & &S &M &L & & &S &M &L\\
    \midrule
    \midrule    
    Cutout~\cite{devries2017improved} &\multirow{5}{*}{\shortstack{ResNet-50\\+\\ImageNet}} &42.9 &32.6 &20.2 & &\multirow{5}{*}[-5pt]{\shortstack{Faster\\R-CNN\\+\\Pascal\\VOC}} &47.8 &65.7 &84.5\\
    Mixup~\cite{zhang2017mixup} & &43.1 &31.7 &19.4 & & &49.1 &68.4 &84.1\\
    CutMix~\cite{kim2020puzzle} & &41.0 &30.5 &18.9 & & &41.4 &67.8 &84.3\\
    PuzzleMix~\cite{kim2020puzzle} & &41.4 &30.1 &18.9 & & &44.1 &63.3 &81.5\\
    SaliencyMix~\cite{uddin2021saliencymix} & &41.9 &29.6 &18.9 & & &47.2 &68.0 &82.8 \\
    \textbf{ContextMix} & &\textbf{40.1} &\textbf{28.6} &\textbf{18.8} & & &\textbf{53.2} &\textbf{68.6} &\textbf{84.2}\\
    \bottomrule
\end{tabular}
\end{center}
\caption{Classification and Detection result on the separate dataset by object size as COCO-style. S, M, and L indicate Small objects, Medium objects, and Large objects, respectively.}
\label{tab:classification_size}
\end{table}

\subsection{Size-wise Classification and Detection} \label{sec:abl_sizewise}
We expect that the sizes of images are generated uniformly by the beta distribution. 
Rather, the small- or medium-sized images are more generated than the large-sized images, as shown in \cref{fig:classification_size}.
We analyze that the clipping operation is affected. The area of the cropping region is decided by beta distribution initially and is generated uniformly. In general, the region is not placed in a clean image. The regions are adjusted by clipping operation for making valid images, which leads to generating small- or medium-sized images. 

As the resize operation leads to the loss of important information for recognizing the object, by making object sizes smaller, one might argue that it could only improve large object recognition performance at the cost of small object recognition capability. Thus, we verify the performance of ContextMix-trained networks on size-wise object recognition on a modified ImageNet validation dataset. The dataset is split using ImageNet object localization information. The separated data consists of small (less than 32$\times$32), medium (from 32$\times$32 to 96$\times$96), and large (greater than 96$\times$96) by object boundary size. And each data quantity is 1,121, 6,756, and 36,371. The remaining 5,743 images consist of mixed sizes and are not appropriate for this experiment. Also, the Pascal VOC dataset for object detection tasks consists in the same manner as COCO-style. \cref{tab:classification_size} shows resize operation does not harm network performance. Rather, ContextMix has more opportunities to learn small- and medium-sized objects and shows better performance in small- and medium-sized object classification and detection.

In this regard, in an experiment exploiting industrial images, our proposed method recorded competitive results, in Top-1 error and F1 scores, to Mixup and showed better performances than Cutout, CutMix, and the saliency-based approaches. Given that the region of interest in industrial images is smaller than the entire image size, our proposed method is available to show improved performances by resized pasting images.
\begin{figure}[t]
    \centering
    \resizebox{1\linewidth}{!}{
    \setlength{\tabcolsep}{1pt}
    \begin{tabular}{ccc}
        \includegraphics[width=1.1\linewidth]{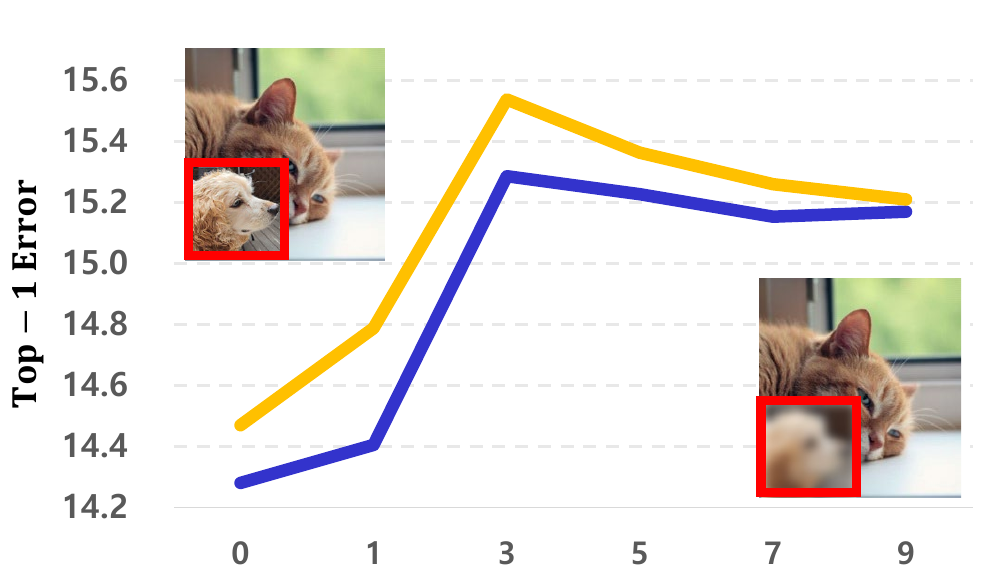}
        & &\includegraphics[width=1.1\linewidth]{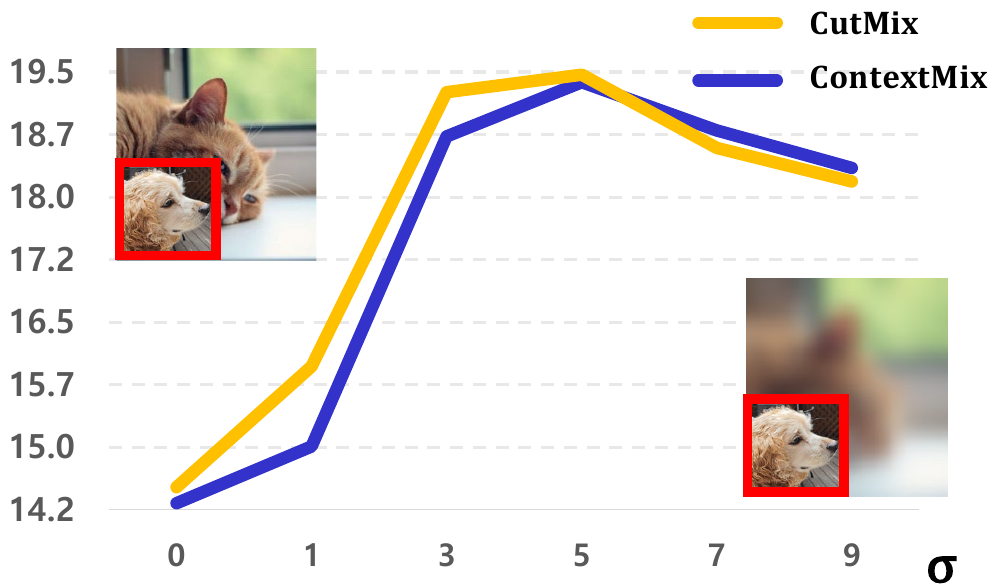}\\
        \fontsize{0.55cm}{0.55cm}\selectfont{\quad(a) Blurring on the Occluded Image}
        & &\fontsize{0.55cm}{0.55cm}\selectfont{\quad\quad(b) Blurring on the Resized Image}\\
    \end{tabular}}
    \caption{Robustness of Gaussian blur($\sigma$) on the occluded image and resized image.}
    \label{fig:gaussian_blur}
\end{figure}
\subsection{Robustness of Gaussian Blur} \label{sec:abl_rbst_blur}
We hypothesize that ContextMix can learn discriminative features by occluded images and context features by resized images. In general, as blur level $\sigma$ increases, the amount of lost information should be increases. The experiments are designed to compare CutMix and ContextMix when increasing the loss of object information.

We verify the first hypothesis, which is ContextMix learns discriminative features by occluded regions. If networks train using blurred occluded images, discriminative features are not utilized to learn, and the performances will degrade. As shown in \cref{fig:gaussian_blur} (a), both CutMix and ContextMix results are similar and are harmed by the blurred operation, then the hypothesis is proved.

Also, we verify the second hypothesis, which is our method learns context features by resized images. If resized or cropped regions don't have sufficient information to recognize objects, increasing the blur level on the regions will not affect the experiment results. And it will show similar results with level 0, such as ContextMix and CutMix. \cref{fig:gaussian_blur} (b) presents that ContextMix is more robust than CutMix and that losing an amount of object information by resizing operation is not a crucial effect for object recognition ability. Instead, it means resized images have more information, including contextual, not only objects.

\subsection{Resize Ratio} \label{sec:abl_resizeratio}
\begin{figure}[!t]
    \centering
    \resizebox{0.9\linewidth}{!}{
    \setlength{\tabcolsep}{1pt}
    \begin{tabular}{c}
        \includegraphics[width=1\linewidth]{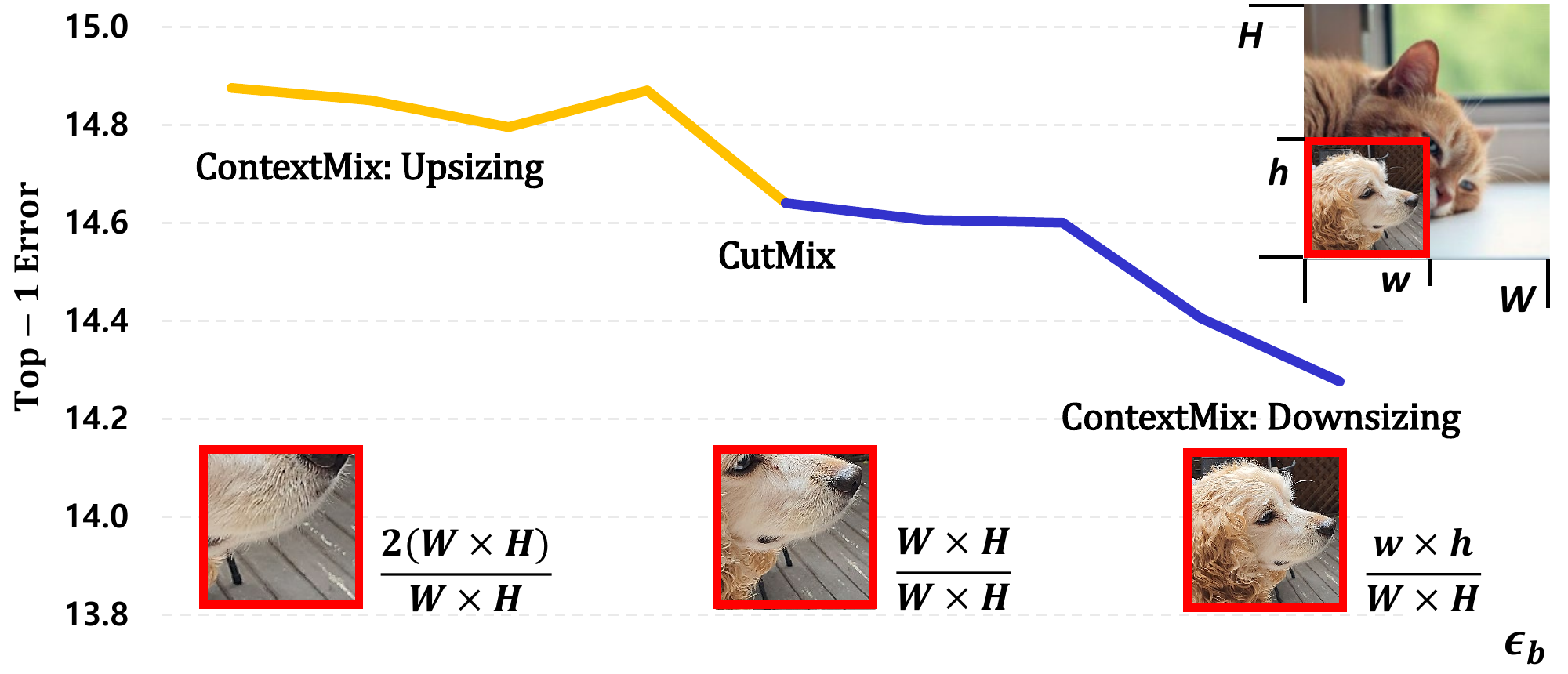}\\
    \end{tabular}}
    \caption{Effects of Resize ratio$(\epsilon_{b})$ from the Upsizing(Left) to the Downsizing(Right) operation of the resized image.}
    \label{fig:graph_resize_ratio}
\end{figure}

In Equation~\ref{eq:3}, we notice that there are various cases of $\mathit{W^{\prime}}$ and $\mathit{H^{\prime}}$. \cref{fig:graph_resize_ratio} shows the effects of three different schemes of resizing ratios. The schemes are CutMix, ContextMix using upsizing images ($\mathit{2W\times 2H}$ $\mathit{\sim}$ $\mathit{W\times H}$), and ContextMix using downsizing ($\mathit{W\times H}$ $\mathit{\sim}$ $\mathit{w\times h}$). CutMix uses the fixed ratio, $\mathit{W\times H}$. ContextMix with downsizing presents the best performance than other schemes. Furthermore, as the downsizing ratio increases until it fits the size of the occluded region, the classification accuracy is increased. It means pasting the downsized images to fit occlusion can preserve the whole image structure, increasing the amount of including object and contextual information. Then it leads to improving object recognition performance.
\begin{table}[t]
\footnotesize
\begin{center}
\setlength{\tabcolsep}{1pt}
\begin{tabular}{p{0.25\linewidth}p{0.45\linewidth}P{0.12\linewidth}P{0.12\linewidth}}
    \toprule
    \multirow{2}{*}[-2pt]{Model}& \multirow{2}{*}[-2pt]{Method}& \multicolumn{2}{c}{Top-1 Error ($\%$)}\\
    \cmidrule{3-4} 
    & &Mean &Std.\\
    \midrule
    \midrule
    \multirow{14}{*}{\shortstack{\fontsize{0.3cm}{0.3cm}\selectfont PyramidNet-200\\($\alpha=240$)}} &Baseline &16.45 &-\\
    &\textbf{ContextMix} &\textbf{14.13} &$\pm$\textbf{~0.20}\\
    \cmidrule{2-4}
    &ContextMix + Center Gaussian &14.60 &$\pm$~0.00\\
    &ContextMix + Fixed Size &14.45 &$\pm$~0.19\\
    &ContextMix + One-hot &15.67 &$\pm$~0.17\\
    &ContextMix + Complete-label &14.94 &$\pm$~0.13\\
    &ContextMix + Scheduled($\uparrow$) &14.91 &$\pm$~0.07\\
    &ContextMix + Scheduled($\downarrow$) &14.71 &$\pm$~0.08\\
    &ContextMix + Square region & 14.59 &$\pm$~0.11\\
    \cmidrule{2-4}
    &ContextMix + Unsharp &14.27 &$\pm$~0.13\\
    &ContextMix + Morphology Erosion &14.55 &$\pm$~0.09\\
    &ContextMix + Morphology Dilation &14.86 &$\pm$~0.15\\
    &ContextMix + Morphology Opening &14.84 &$\pm$~0.17\\
    &ContextMix + Morphology Closing &14.64 &$\pm$~0.10\\
    \bottomrule
\end{tabular}
\end{center}
\caption{ContextMix variants results on CIFAR-100.}
\label{tab:ablation_variant_cifar100_ContextMix}
\end{table}
\subsection{ContextMix variants} \label{sec:abl_variant}
Table~\ref{tab:ablation_variant_cifar100_ContextMix} shows the performances of the ContextMix variants. We briefly describe the experimental schemes. ‘Center Gaussian’ method fixes the bounding box starting point ($\mathit{r_{xs}, r_{ys}}$) to the center of images, with ranges $\mathit{w, h}$ of cropped regions using Gaussian distribution. ‘Fixed-size' method fixes the cropped region's size at 75$\%$. ‘One-hot' is set to the label that occupies a larger space on a newly generated sample, rather than using label smoothing. ‘Complete-label' set to the new label $\mathit{\tilde{y}}$ with equivalent $y_A$ and $y_B$, in a rate 5:5. `Scheduled($\uparrow$)' increases the probability of using ContextMix in training from 0 to 1. `Scheduled($\downarrow$)' decreases the probability of applying ContextMix in training from 1 to 0. ContextMix can generate resized areas by different ratios of widths and heights. `Square region' fixes the cropped regions to squares and does resize operation. To explore the robustness of ContextMix distortions, we experiment with adjusting various image distortions, such as sharpening and various morphology techniques. From the results, we verify that occluded images and resized are not contradictions; instead are cooperative relations.
\begin{table}[t]
\footnotesize
\begin{center}
\setlength{\tabcolsep}{1pt}
\renewcommand{\arraystretch}{1.}
\begin{tabular}{P{0.40\columnwidth}p{0.15\columnwidth}P{0.15\columnwidth}P{0.22\columnwidth}}
    \toprule
    &Method &Region &Top-1 Error~($\%$)\\
    \midrule
    \midrule
    \multirow{4}{*}{\includegraphics[width=0.23\columnwidth]{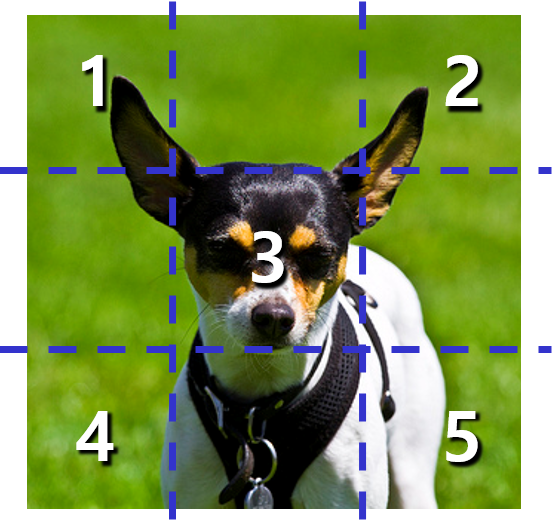}} &ContextMix &- &21.03\\
    \cmidrule{2-4}
    &\multirow{4}{*}{\shortstack{Fixed-\\ContextMix}} &1 &22.92\\
    & &2 &22.98\\
    & &3 &23.08\\
    & &4 &23.20\\
    & &5 &23.02\\
    \bottomrule
\end{tabular}
\end{center}
\caption{Effects of Fixed region and Resize ratio: Classification results with ResNet-50 on ImageNet validation.}
\label{tab:fixed_sample}
\end{table}
\subsection{Fixed resizing ratio and pasting region} \label{sec:abl_fixed}
Our proposed method utilizes random location pasting and random resizing. As shown in Table~\ref{tab:fixed_sample} an example image, we conduct contrary experiments, Fixed-ContextMix. The inputs of experiments are not random location and random resizing; instead are divided into nine areas, the resize ratio is fixed to the size of the divided area, and the pasting location is set to the five regions from each evaluation, respectively. 
\cite{zhang2018adversarial}, \cite{Kim_2017_ICCV} and \cite{wei2017object} proposed AE and confirmed that the dropout-ed regions do not hinder object recognition performance but rather acquire opportunities for training diverse and overall object features.
Table~\ref{tab:fixed_sample} re-validates that random location and random resizing ratio are helped to learn details and more discriminative features of whole objects. And we verify that the randomness of our proposed method leads to the performance of various tasks.

\subsection{Limitation} \label{sec:abl_limit}
Through the random selection of cropped and pasted regions across various iterations and epochs, we have observed that these regions can offer opportunities to train features with increased robustness. Nevertheless, there are still untapped possibilities to harness information from these cropped regions.
It is worth noting that our experimentation involved datasets like CIFAR, ImageNet, and Pascal VOC, all of which consist of objects with varying sizes within the human visual spectrum. These datasets were chosen to represent common scenarios in computer vision. Notably, our industrial dataset, which we used for training and evaluation, differs in its focus. It did not prioritize recognizing small objects, such as particles or lint, but rather considered manufactured objects within a controlled cleanroom environment.
To address these specific challenges, we have concrete plans to expand our research. One key aspect of our future work is centered around improving the recognition of tiny objects within large-scale images. We acknowledge that this area presents unique difficulties, and we aim to develop strategies and models to tackle these challenges effectively. Additionally, we intend to further explore and exploit the valuable features that can be extracted from cropped regions in both small and large objects to enhance the overall performance of our computer vision systems.

\section{Conclusion} \label{sec:conclusion}
In this paper, we have introduced a simple yet powerful data augmentation technique, ContextMix. 
The proposed method generates unseen data using the entire resized image, leading to improved classification performance.
We have demonstrated the effectiveness of the proposed ContextMix in object detection and semantic segmentation tasks on the PASCAL VOC dataset with pre-trained weights. In addition, ContextMix has exhibited higher robustness compared to other augmentation methods when subjected to FGSM white-box attacks, natural adversarial datasets, and ImageNet-A, respectively. These experimental results indicate the broad applicability of the proposed method across various computer vision tasks with a high level of robustness.

To show the effectiveness and applicability of the proposed method, we evaluated the proposed method on the real-world industrial dataset for the passive component inspection provided by SEMCO.
The proposed method shows improvement up to 2$\%$ in terms of Top-1 accuracy and macro F1 score compared to the state-of-the-art methods on the real-world industrial dataset, showcasing its superiority in a practical setting.
The key modification of resizing the image played a crucial role in improving performances across different tasks. We performed an in-depth analysis to understand the underlying reasons for this improvement.
Furthermore, our approach is designed to be straightforward and intuitive, making it easily understandable and accessible for managers or experts who work in manufacturing lines.

From CAM and weakly supervised object localization task results, we observed that ContextMix generated predicted object regions that were more encompassing compared to CutMix and PuzzleMix.
This observation suggests that ContextMix not only trains the model on objects but also captures contextual information.
Various ablation studies confirmed that the resizing operation does not lead to the loss of crucial information but instead improves the recognition capability.
\begin{figure}[t]
    \centering
    \resizebox{1.\linewidth}{!}{
    \setlength{\tabcolsep}{5pt}
    \begin{tabular}{ccc}
        \includegraphics[width=0.85\linewidth]{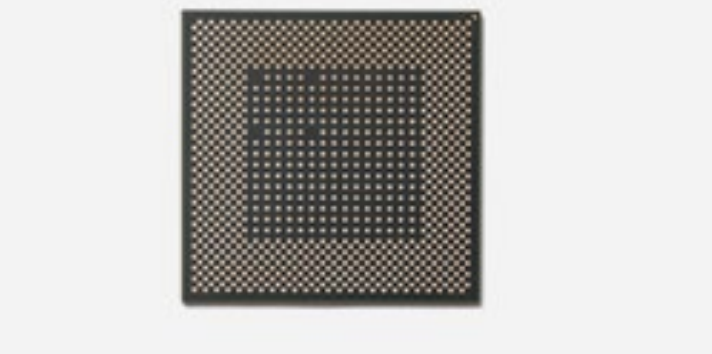}  &\includegraphics[width=0.85\linewidth]{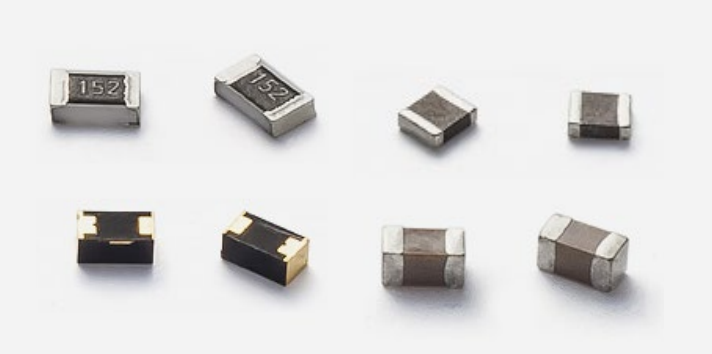}
        &\includegraphics[width=0.85\linewidth]{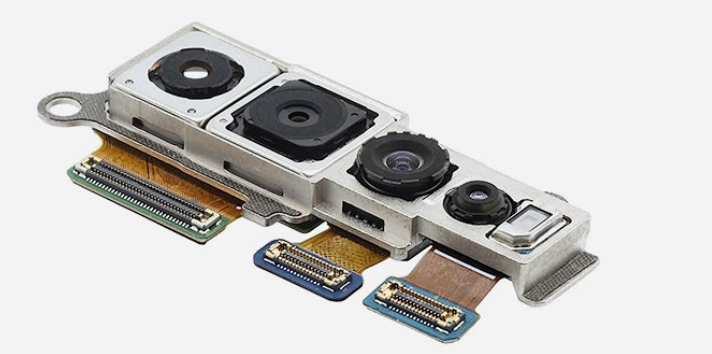}\\
        \fontsize{0.7cm}{0.7cm}\selectfont{(a) Package substrate} &\fontsize{0.7cm}{0.7cm}\selectfont{(b) Passive component} &\fontsize{0.7cm}{0.7cm}\selectfont{(c) Camera module}\\
    \end{tabular}}
    \caption{The devices are produced by Samsung electro-mechanics and essentially required the auto optical inspection system for efficiency in modern industrial environments.}
    \label{fig:semproducts}
\end{figure}

Additionally, we presented the contradiction of the saliency-based method that the salient region can be the background, not the object. 
ContextMix overcomes this limitation by generating images containing object information along with contextual information, without incurring additional computational costs for acquiring salient regions. The superiority of ContextMix over other data augmentation methods has been demonstrated across different tasks, datasets, and network architectures.
In future work, we envision its application in inspection machines for other products of SEMCO,
such as the products shown in Figure~\ref{fig:semproducts}.
Furthermore, we plan to extend ContextMix to image-to-text translation and natural language processing tasks, opening up new possibilities for leveraging this augmentation technique in various domains.

\section*{Acknowledgement}
This work was funded by Samsung Electro-Mechanics and was partially supported by Carl-Zeiss Stiftung under the Sustainable Embedded AI project (P2021-02-009).

\bibliography{article}


\end{document}